\documentclass[sigconf]{acmart}
\renewcommand\footnotetextcopyrightpermission[1]{}
\usepackage{xspace}
\usepackage{amsmath}
\usepackage{epsfig}
\usepackage{enumerate}
\usepackage{makecell}
\usepackage{multirow}
\usepackage{epstopdf}
\usepackage{graphicx}
\usepackage{amsfonts}
\usepackage{tikz}
\usepackage{algorithm}
\usepackage{algorithmic}
\usepackage{enumitem}
\usepackage{tabularx}

\usepackage{bm,amsmath,amssymb}
\usepackage{xspace}
\usepackage{graphicx}
\usepackage{color}
\usepackage{natbib}

\usepackage{colortbl}
\definecolor{light_gray}{RGB}{170,170,170}

\usepackage[utf8]{inputenc} 
\usepackage[T1]{fontenc}    
\usepackage{hyperref}       
\usepackage{url}            
\usepackage{booktabs}       
\usepackage{amsfonts}       
\usepackage{nicefrac}       
\usepackage{microtype}      
\usepackage{longtable,lscape}

\usepackage{subcaption}

\usepackage{wrapfig}

\newcommand{\hide}[1]{}
\DeclareMathOperator*{\argmax}{argmax}
\DeclareMathOperator*{\argmin}{argmin}

\usepackage{enumitem}

\usepackage{bm}

\newcommand{\mname}{\texttt{CHEER}\xspace}
\newcommand{\tname}{\mname}

\copyrightyear{2020}
\acmYear{2020}
\setcopyright{rightsretained}

\DeclareMathOperator*{\minimize}{\text{minimize}}

\begin{document}

\title{\tname: 	Rich Model Helps Poor Model via Knowledge Infusion}

\author{Cao~Xiao}
\authornote{Equal contribution}
\affiliation{%
\institution{Analytics Center of Excellence, IQVIA}
\city{Cambridge, MA}
\country{USA}}
\email{cao.xiao@iqvia.com}

\author{Trong Nghia~Hoang}
\authornotemark[1]
\affiliation{%
\institution{MIT-IBM Watson AI Lab}
\city{Cambridge, MA}
\country{USA}}
\email{nghiaht@ibm.com}

\author{Shenda~Hong}
\authornotemark[1]
\affiliation{%
\institution{Department
of Computer Science and Engineering, Georgia Institute of Technology}
\city{Atlanta, GA}
\country{USA}}
\email{sdhong1503@gmail.com}

\author{Tengfei~Ma}
\affiliation{%
\institution{IBM Research}
\city{Yorktown Heights, NY}
\country{USA}}
\email{Tengfei.Ma1@ibm.com}

\author{Jimeng~Sun}
\affiliation{%
\institution{Department of Computer Science, University of Illinois at Urbana-Champaign}
\city{Urbana, IL}
\country{USA}}
\email{jimeng@illinois.edu}

\begin{abstract}
There is a growing interest in applying deep learning (DL) to healthcare, driven by the availability of data with multiple feature channels in \emph{rich-data} environments (e.g., intensive care units). However, in many other practical situations, we can only access data with much fewer feature channels in a \emph{poor-data} environments (e.g., at home), which often results in predictive models with poor performance. How can we boost the performance of models learned from such \emph{poor-data} environment by leveraging knowledge extracted from existing models trained using \emph{rich data} in a related environment? To address this question, we develop a knowledge infusion framework named \tname that can succinctly summarize such \emph{rich model} into transferable representations, which can be incorporated into the \emph{poor model} to improve its performance. The infused model is analyzed theoretically and evaluated empirically on several datasets. Our empirical results showed that \tname outperformed baselines by $5.60\%$ to $46.80\%$ in terms of the macro-F1 score on multiple physiological datasets.
\end{abstract}


\maketitle
\pagestyle{plain}

\section{Introduction}
\label{intro}
In \emph{rich-data} environments with strong observation capabilities, data often come with rich representations that encompass multiple channels of features. 
For example, multiple leads of Electrocardiogram (ECG) signals in hospital used for diagnosing heart diseases are measured in intensive care units (ICU), of which each lead is considered a feature channel. The availability of such rich-data environment has thus sparked strong interest in applying deep learning (DL) for predictive health analytics as DL models built on data with multi-channel features have demonstrated promising results in healthcare~\cite{doi:10.1093/jamia/ocy068}. However, in many practical scenarios, such rich data are often private and not accessible due to privacy concern. Thus, we often have to develop DL models on lower quality data comprising fewer feature channels, which were collected from \emph{poor-data} environments with limited observation capabilities (e.g., home monitoring devices which provide only a single channel of feature). Inevitably, the performance of state-of-the-art DL models, which are fueled by the abundance and richness of data, becomes much less impressive in such poor-data environments~\cite{DBLP:journals/corr/abs-1708-04347}.\\

\noindent To alleviate this issue, we hypothesize that learning patterns consolidated by DL models trained in one environment often encode information that can be transferred to related environments. For example, a heart disease detection model trained on \emph{rich-data} from 12 ECG channels in a hospital will likely carry pertinent information that can help improve a similar model trained on \emph{poor-data} from a single ECG channel collected by a wearable device due to the correlation between their data. Motivated by this intuition, we further postulate that given access to a prior model trained on \emph{rich-data}, the performance of a DL model built on a related \emph{poor-data} can be improved if we can extract transferable information from the \emph{rich} model and infuse them into the \emph{poor} model. This is related to  deep transfer learning and knowledge distillation but with a new setup that has not been addressed before, as elaborated in Section~\ref{sec:related} below.\\ 

\noindent In this work, we propose a knowledge infusion framework, named \tname, to address the aforementioned challenges. In particular, \tname aims to effectively transfer domain-invariant knowledge consolidated from a \emph{rich} model with high-quality data demand to a \emph{poor} model with low data demand and model complexity, which is more suitable for deployment in \emph{poor-data} settings. We also demonstrate empirically that \tname helps bridge the performance gap between DL models applied in \emph{rich}- and \emph{poor}-data settings. Specifically, we have made the following key contributions:\\

\noindent {\bf 1.} We develop a transferable representation that summarizes the \emph{rich model} and then infuses the summarized knowledge effectively into the \emph{poor model} (Section~\ref{representation}). The representation can be applied to a wide range of existing DL models.\\

\noindent {\bf 2.} We perform theoretical analysis  to demonstrate the efficiency of  knowledge infusion mechanism of \tname. Our theoretical results show that under practical learning configurations and mild assumptions, the \emph{poor model}'s prediction will agree with that of the \emph{rich model} with high probability (Section~\ref{theory}).\\

\noindent {\bf 3.} Finally, we also conduct extensive empirical studies to demonstrate the efficiency of \tname on several healthcare datasets. Our results show that \tname outperformed the second best approach (knowledge distillation) and the baseline without knowledge infusion by $5.60$\% and $46.80$\%, respectively, in terms of macro-F1 score and demonstrated more robust performance (Section~\ref{experiment}).

\section{Related Works}
\label{sec:related}
\noindent {\bf Deep Transfer Learning}: 
Most existing deep transfer learning methods transfer knowledge across domains while assuming the target and source models have equivalent modeling and/or data representation capacities. For example, deep domain adaptation have focused mainly on learning domain-invariant representations between very specific domains (e.g., image data) on  ~\cite{Glorot:2011:DAL:3104482.3104547, Chen:2012:MDA:3042573.3042781,Ganin:2016:DTN:2946645.2946704,Zhou2016BiTransferringDN,Bousmalis:2016:DSN:3157096.3157135,LongC0J15,HuangZL18,RozantsevSF19,XuHZT18}. Furthermore, this can only be achieved by training both models jointly on source and target domain data.\\ 

\noindent More recently, another type of deep transfer learning~\cite{Zagoruyko2017AT} has been developed to transfer only the attention mechanism~\cite{bahdanau2014neural} from complex to shallow neural network to boost its performance. Both source and target models, however, need to be jointly trained on the same dataset. In our setting, since source and target datasets are not available at the same time and that the target model often has to adopt representations with significantly less modeling capacity to be compatible with the \emph{poor-data} domain with weak observation capabilities.\\ 

\noindent {\bf Knowledge Distillation}: Knowledge distillation~\cite{hinton2015distilling} or mimic learning ~\cite{ba2014deep} aim to transfer the predictive power from a high-capacity but expensive DL model to a simpler model such as shallow neural networks for ease of deployment~\cite{sau2016deep,radosavovic2017data,yim2017gift,lopez2015unifying}. This can usually be achieved via training simple models on soft labels learned from high-capacity models, which, however, assume that both models operate on the same domain and have access to the same data or at least datasets with similar qualities. In our setting, we only have access to low-quality data with \emph{poor} feature representation, and an additional set of limited \emph{paired} data that include both \emph{rich} and \emph{poor} representations (e.g., high-quality ICU data and lower-quality health-monitoring information from personal devices) of the same object.\\

\noindent {\bf Domain Adaptation}: There also exists another body of non-deep learning transfer paradigms that were often referred to as domain adaption. This however often include methods that not only assume access domain-specific \cite{WeiKG19,WangQWLGZHZCZ19,LuoCTSLCY18,PengTXWPH18,TangWWGDGC18} and/or model-specific knowledge of the domains being adapted~\cite{PanTKY09,PanTKY11,yao2019heterogeneous,JiangGLLCH19,LiXXDG18,LuoWLT19,SegevHMCE17,XuPXWLMS17}, but are also not applicable to deep learning models~\cite{GhifaryBKZ17,WuWZTXYH17} with arbitrary architecture as addressed in our work.\\

\noindent In particular, our method does not impose any specific assumption on the data domain and the deep learning model of interest. We recognize that our method is only demonstrated on deep model (with arbitrary architecture) in this research but our formulation can be straightforwardly extended to non-deep model as well. We omit such detail in the current manuscript to keep the focus on deep models which are of greater interest in healthcare context due to their expressive representation in modeling multi-channel data.

\section{The \mname Method}
\label{formulation}

\begin{table}
\centering
\caption{Notations used in \mname.}
\begin{tabularx}{\columnwidth}{l|l}
\toprule[1pt]
\bf \hspace{-1.5mm}Notation & \bf Definition \\
\hline 
\hspace{-1.5mm}$\mathcal{H}_r \triangleq \{(\mathbf{x}_i^r, y_i^r)\}_{i=1}^n$; $\mathcal{T}\left(y\ |\ \mathbf{x}^r\right)$ 
 & rich data; rich model \\
\hspace{-1.5mm}$\mathcal{H}_p \triangleq \{(\mathbf{x}_i^p,y_i^p)\}_{i=1}^m$; $\mathcal{S}\left(y\ |\ \mathbf{x}^p\right)$ & poor data; poor model \\
\hspace{-1.5mm}$\mathcal{H}_o \triangleq \{(\mathbf{x}_i^r, \mathbf{x}_i^p, y_i)\}_{i=1}^{k}$ & paired data \\
\hspace{-1.5mm}$\mathbf{Q}_r \triangleq [q_r^1 \ldots q_r^{l_r}] \in \mathbb{R}^{d\times l_r}$ & domain-specific features \\
\hspace{-1.5mm}$\mathbf{A}_r(\mathbf{x}^r) \triangleq [a_r^{(1)}(\mathbf{x}^r) \ldots a_r^{(d)}(\mathbf{x}^r)]$ & feature scoring functions\\
\hspace{-1.5mm}$\mathbf{O}_r \triangleq \mathbf{O}_r(\mathbf{Q}_r^\top\mathbf{A}_r(\mathbf{x}^r))$ & feature aggregation component\\
\bottomrule[1pt]
\end{tabularx}
\label{tab:symbol}
\end{table}

\subsection{Data and Problem Definition}
\label{notation} 

{\bf Rich and Poor Datasets.} Let $\mathcal{H}_r \triangleq \{(\mathbf{x}_i^r, y_i^r)\}_{i=1}^n$ and 
$\mathcal{H}_p \triangleq \{(\mathbf{x}_i^p,y_i^p)\}_{i=1}^m$ denote the \emph{rich} and \emph{poor} datasets, respectively. The subscript $i$ indexes the $i$-th data point (e.g., the $i$-th patient in healthcare applications), which contains input feature vector $\mathbf{x}_i^r$ or $\mathbf{x}_i^p$ and output target $y_i^r$ or $y_i^p$ of the rich or poor datasets. The \emph{rich} and \emph{poor} input features $\mathbf{x}_i^r\in\mathbb{R}^r$ and $\mathbf{x}_i^p\in\mathbb{R}^p$ are $r$- and $p$-dimensional vectors with $p \ll r$, respectively. The output targets, $y_i^r$ and $y_i^p \in \{1\ldots c\}$, are categorical variables. The input features of these datasets (i.e., $\mathbf{x}_i^r$ and $\mathbf{x}_i^p$) are non-overlapping as they are assumed to be collected from different channels of data (i.e., different {\bf data modalities}). In the remaining of this paper, we will use {\bf data channel} and {\bf data modality} interchangeably.\\

\noindent For example, the rich data can be the physiological data from ICU (e.g., vital signs, continuous blood pressure and electrocardiography) or temporal event sequences such as electronic health records with discrete medical codes, while the poor data are collected from personal wearable devices. The target can be the mortality status of those patients, onset of heart diseases and etc. Note that these raw data are not necessarily plain feature vectors. They can be arbitrary rich features such as time series, images and text data. We will present one detailed implementation using time series data in Section~\ref{sec:dnn}.\\ 

\noindent {\bf Input Features.} We (implicitly) assume that the raw data of interest comprises (says, $p$ or $r$) multiple sensory channels, each of which can be represented by or embedded\footnote{We embed these channel jointly rather than separately to capture their latent correlation.} into a particular feature signal (i.e., {\bf one feature per channel}). This results in an embedded feature vector of size $p$ or $r$ (per data point), respectively. In a different practice, a single channel may be encoded by multiple latent features and our method will still be applicable. In this paper, however, we will {\bf assume one embedded feature per channel} to remain close to the standard setting of our healthcare scenario, which is detailed below.\\

{\noindent\bf Paired Dataset.} To leverage both rich and poor datasets, we need a small amount of {\bf paired data} to learn the relationships between them, which is denoted as $\mathcal{H}_o \triangleq \{(\mathbf{x}_i^r, \mathbf{x}_i^p, y_i)\}_{i=1}^{k}$. Note that the paired dataset contains both \emph{rich} and \emph{poor} input features, i.e. $\mathbf{x}_i^r$ and $\mathbf{x}_i^p$, of the same subjects (hence, sharing the same target $y_i$).\\ 

\noindent Concretely, this means a concatenated input $\mathbf{x}_i^o = [\mathbf{x}_i^r, \mathbf{x}_i^p]$ of the paired dataset has $o = p + r$ features where the first $r$ features are collected from $r$ rich channels (with highly accurate observation capability) while the remaining $p$ features are collected from $p$ poor channels (with significantly more noisy observations). We note that our method and analysis also apply to settings where $\mathbf{x}_i^p \subseteq \mathbf{x}_i^r$. In such cases, $\mathbf{x}_i^o = \mathbf{x}_i^r$ and $o = r$ (though the number of data point $i$ for which $\mathbf{x}_i^r$ is accessible as paired data is much less than the number of those with accessible $\mathbf{x}_i^p$). To avoid confusion, however, we will proceed with the implicit assumption that there is no feature overlapping between poor and rich datasets in the remaining of this paper.\\

\noindent For example, the paired dataset may comprise of rich data from ICU ($\mathbf{x}^p_i$) and poor data from wearable sensors ($\mathbf{x}^r_i$), which are extracted from the same patient $i$. The paired dataset often contains much fewer data points (i.e., patients) than the rich and poor datasets themselves, and cannot be used alone to train a prediction model with high quality. \\

{\noindent\bf Problem Definition.} Given (1) a poor dataset $\mathcal{H}_p$ collected from a particular patient cohort of interest, (2) a paired dataset $\mathcal{H}_o$ collected from a limited sample of patients, and (3) a \emph{rich} model $\mathcal{T}(y|\mathbf{x}^r)$ which were pre-trained on private (rich) data of the same patient cohort, we are interested in learning a model $\mathcal{S}(y|\mathbf{x}^p)$ using both $\mathcal{H}_p$, $\mathcal{H}_o$ and $\mathcal{T}(y|\mathbf{x}^r)$, which can perform better than a vanilla model $\mathcal{D}(y|\mathbf{x}^p)$ generated using only $\mathcal{H}_p$ or $\mathcal{H}_o$.\\

\noindent {\bf Challenges.} This requires the ability to transfer the learned knowledge from $\mathcal{T}(y|\mathbf{x}^r)$ to improve the prediction quality of $\mathcal{S}(y|\mathbf{x}^p)$. This is however a highly non-trivial task because (a) $\mathcal{T}(y|\mathbf{x}^r)$ only generates meaningful prediction if we can provide input from rich data channels, (b) its training data is private and cannot be accessed to enable knowledge distillation and/or domain adaptation, and (c) the paired data is limited and cannot be used alone to build an accurate prediction model.\\

\begin{figure}[ht]
\centering
\hspace{-1mm}\includegraphics[width=1.01\linewidth]{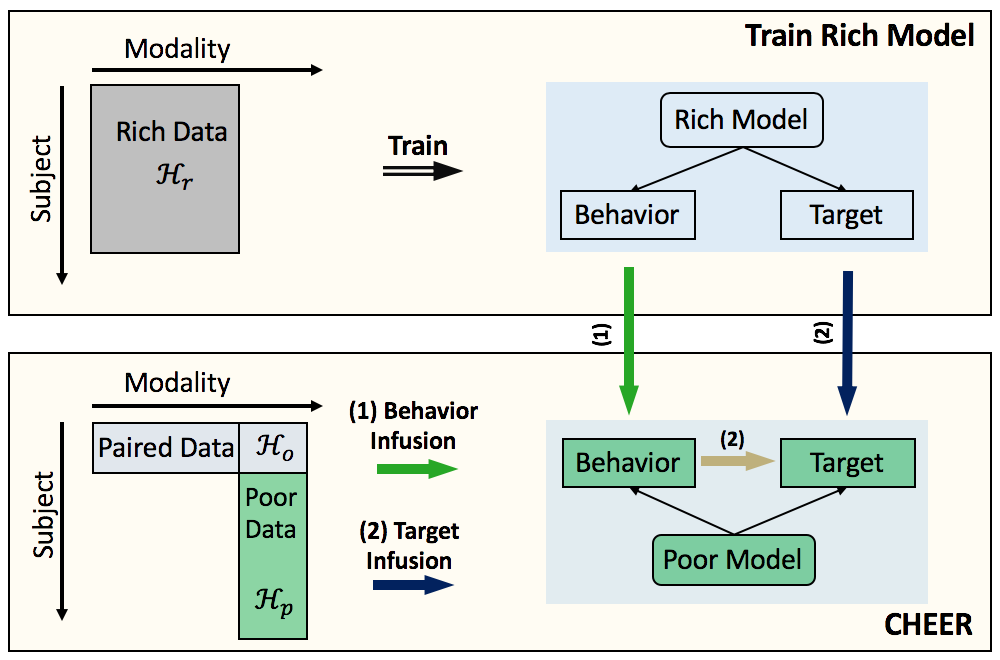}
\caption{\tname: (a) a \emph{rich} model was first built using rich multi-modal or multi-channel data; 
(b) the behaviors of \emph{rich} model are then infused into the poor model using paired data (i.e., behavior infusion); 
and (c) the \emph{poor} model is trained to fit both the rich model's predictions on paired data and its \emph{poor} dataset (i.e., target infusion).}
\label{fig:framework}
\end{figure}

\noindent {\bf Solution Sketch.} Combining these sources of information coherently to generate a useful prediction model on the patient cohort of interest is therefore a challenging task which has not been investigated before. To address this challenge, the idea is to align both rich and poor models using a transferable representation described in Section~\ref{representation}. This representation in turn helps infuse knowledge from the rich model into the poor model, thus improving its performance. The overall structure of \tname is shown in Figure ~\ref{fig:framework}. The notations are summarized in Table~\ref{tab:symbol}.

\subsection{Learning Transferable Rich Model}
\label{representation}

In our knowledge infusion task, the \emph{rich} model is assumed to be trained in advance using the rich dataset $\mathcal{H}_r \triangleq \{(\mathbf{x}_i^r, y_i^r)\}_{i=1}^n$. The rich dataset is, however, not accessible and we only have access to the \emph{rich} model. The knowledge infusion task aims to consolidate the knowledge acquired by the rich model and infuse it with a simpler model (i.e., the \emph{poor} model).\\

\noindent {\bf Transferable Representation.} To characterize a DL model, we first describe the building blocks and then discuss how they would interact to generate the final prediction scores. In particular, let $\mathbf{Q}_r(\mathbf{x}^r)$, $\mathbf{A}_r(\mathbf{x}^r)$ and $\mathbf{O}_r$ denote the building blocks, namely  \emph{Feature Extraction}, \emph{Feature Scoring} and \emph{Feature Aggregation}, respectively. 
Intuitively, the \emph{Feature Extraction} first transforms raw input feature $\mathbf{x}^r$ into a vector of high-level features $\mathbf{Q}_r(\mathbf{x}^r)$, whose importance are then scored by the \emph{Feature Scoring} function $\mathbf{A}_r(\mathbf{x}^r)$. The high-level features $\mathbf{Q}_r(\mathbf{x}^r)$ are combined first via a linear transformation $\mathbf{Q}_r(\mathbf{x}^r)^\top\mathbf{A}_r(\mathbf{x}^r)$ that focuses the model's attention on important features. The results are translated into a vector of final predictive probabilities via the \emph{Feature Aggregation} function $\mathbf{O}_r(\mathbf{Q}_r(\mathbf{x}^r)^\top\mathbf{A}_r(\mathbf{x}^r))$, which implements a non-linear transformation. Mathematically, the above workflow can be succinctly characterized using the following conditional probability distributions: 
\begin{eqnarray}
\mathcal{T}\left(y\ |\ \mathbf{x}^r\right) &\triangleq& \mathrm{Pr}\Big(y \ |\  \mathbf{Q}_r^\top\left(\mathbf{x}^r\right)\mathbf{A}_r\left(\mathbf{x}^r\right); \mathbf{O}_r\Big) \ .
\end{eqnarray}
We will describe these building blocks in more details next.\\

\noindent\textbf{Feature Extraction.} Dealing with complex input data such as time series, images and text, it is common to derive more effective features instead of directly using the raw input $\mathbf{x}^r$. The extracted features are denoted as $\mathbf{Q}_r(\mathbf{x}_r) \triangleq [q_r^1(\mathbf{x}^r) \ldots q_r^{l_r}(\mathbf{x}^r)] \in \mathbb{R}^{d\times l_r}$ where $q_r^i(\mathbf{x}^r) \in \mathbb{R}^d$ is a $d$-dimensional feature vector extracted by the $i$-th feature extractor from the raw input $\mathbf{x}^r$. Each feature extractor is applied to a separate segment of the time series input (defined later in Section~\ref{sec:dnn}). To avoid cluttering the notations, we shorten $\mathbf{Q}_r(\mathbf{x}^r)$ as $\mathbf{Q}_r$.\\

\noindent\textbf{Feature Scoring.} 
Since the extracted features are of various importance to each subject, they are combined via weights specific to each subject. More formally, the extracted features $\mathbf{Q}_r(\mathbf{x}^r)$ of the \emph{rich model} are combined via $\mathbf{Q}_r^\top(\mathbf{x}^r)\mathbf{A}_r(\mathbf{x}^r)$ using subject-specific weight vector $\mathbf{A}_r(\mathbf{x}^r) \triangleq [a_r^{(1)}(\mathbf{x}^r)\ldots a_r^{(d)}(\mathbf{x}^r)]\in  \mathbb{R}^d$.\\

\noindent Essentially, each weight component $a_r^{(i)}(\mathbf{x}^r)$ maps from the raw input feature $\mathbf{x}^r$ to the important score of its $i^{\mathrm{th}}$ extracted feature. For each dimension $i$, $a_r^{(i)}(\mathbf{x}^r) \triangleq a_r^{(i)}(\mathbf{x}^r; \boldsymbol{\omega}_r^{(i)})$ parameterized by a set of parameters $\boldsymbol{\omega}_r^{(i)}$, which are learned using the \emph{rich} dataset.\\

\noindent\textbf{Feature Aggregation.} The \emph{feature aggregation}  implements a nonlinear transformation $\mathbf{O}_r$ (e.g., a feed-forward layer) that maps the combined features into final predictive scores. The input to this component is the linearly combined feature $\mathbf{Q}_r(\mathbf{x}^r)^\top\mathbf{A}_r(\mathbf{x}^r)$ and the output is a vector of logistic inputs, 
\begin{eqnarray}
\mathbf{r}\left(\mathbf{x}^r\right) \ \ \triangleq\ \  \left[r_1 \ldots r_c\right] \ \ =\ \ \mathbf{O}_r\left(\mathbf{Q}_r\left(\mathbf{x}^r\right)^\top\mathbf{A}_r\left(\mathbf{x}^r\right)\right) \ ,
\end{eqnarray}
which is subsequently passed through the softmax function to compute the predictive probability for each candidate label,
\begin{eqnarray}
\mathcal{T}\left(y = j|\mathbf{x}^r_i\right) &\triangleq& \exp\left(r_j\right) \Bigg/ \left(\sum_{\kappa = 1}^c \exp\left(r_{\kappa}\right)\right) \ .
\end{eqnarray}

\subsection{A DNN Implementation of Rich Model}\label{sec:dnn}

\begin{figure*}[ht]
\centering
\hspace{-3mm}\includegraphics[width=\linewidth]{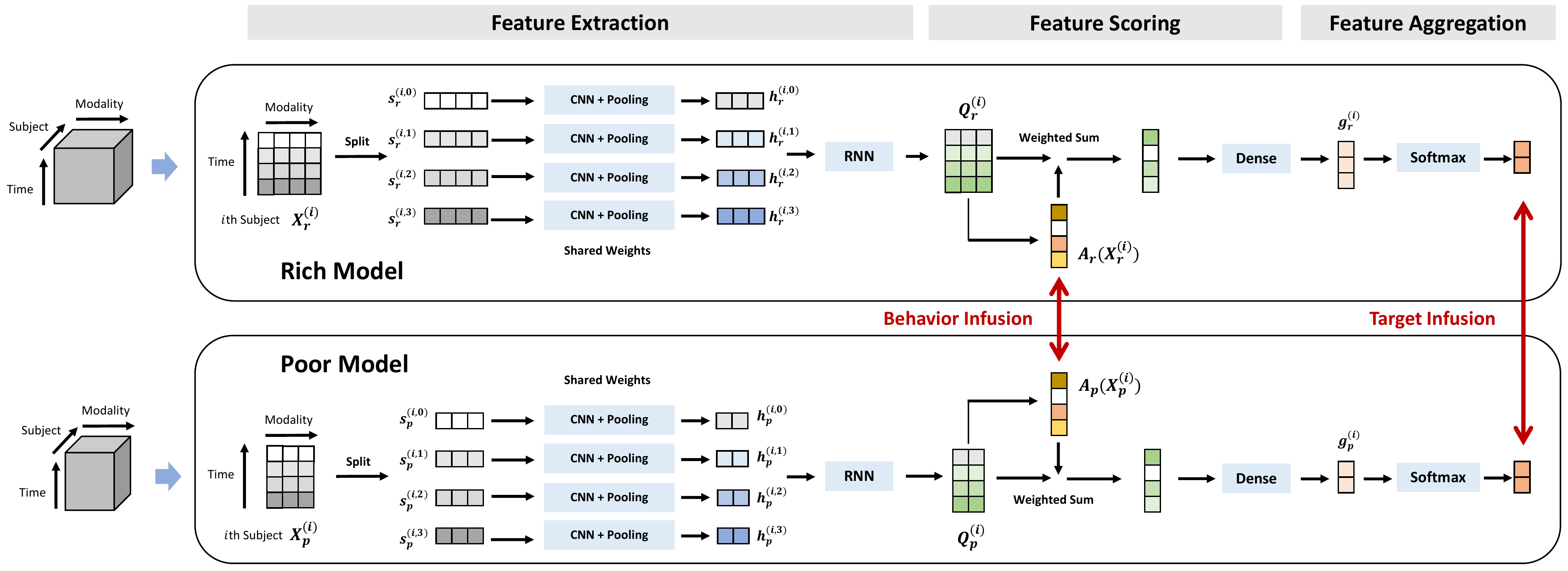}
\caption{The DNN Implementation of \tname. }
\label{fig:dnn}
\end{figure*}

This section describes an instantiation of the aforementioned abstract building blocks using a popular DNN architecture with self-attention mechanism~\cite{lin2017structured} for modeling multivariate time series~\cite{choi2016convolutional}:\\

\noindent{\bf Raw Features.} Raw data  from rich data environment often consist of multivariate time series such as physiological signals collected from hospital or temporal event sequences such as electronic health records (EHR) with discrete medical codes. In the following, we consider the raw feature input $\mathbf{x}^r_i$ as continuous monitoring data (e.g., blood pressure measures) for illustration purpose. \\

\noindent{\bf Feature Extraction.}  To handle such continuous time series, we extract a set of domain-specific features using CNN and RNN models. More specifically, we splits the raw time series $\mathbf{x}^r_i$ into $l_r$ non-overlapping segments of equal length.\\

\noindent That is, $\mathbf{x}^r_i \triangleq \left(\mathbf{s}^r_{i,m}\right)$
where $m = 1 \ldots l_r$ and $\mathbf{s}^r_{i,m} \in \mathbb{R}^{D_r}$ such that $D_r \times l_r = r$ with $r$ denotes the number of features of the rich data. Then, we apply stacked 1-D convolutional neural networks ($\mathbb{CNN}_r$) with mean pooling ($\mathbb{P}_r$) on each segment, i.e. 
\begin{eqnarray}
\mathbf{h}^r_{i,m} &\triangleq& \mathbb{P}_r\left(\mathbb{CNN}_r\left(\mathbf{s}^r_{i,m}\right)\right)
\end{eqnarray}
where $\mathbf{h}^r_{i,m} \in \mathbb{R}^{k_r}$, and $k_r$ denotes the number of filters of the CNN components of the rich model. After that, we place a recurrent neural network ($\mathbb{RNN}_r$) across the output segments of the previous CNN and Pooling layers:
\begin{eqnarray}
\mathbf{q}^r_{i,m} &\triangleq& \mathbb{RNN}_r\left(\mathbf{q}^r_{i,m - 1}, \mathbf{h}^r_{i,m}\right) \ \ \in\ \ \mathbb{R}^d \ ,
\end{eqnarray}

\noindent The output segments of the RNN layer are then concatenated to generate the feature matrix,
\begin{eqnarray}
\mathbf{Q}_r^{(i)} &\triangleq& \left[\mathbf{q}^r_{i,1} \ldots \mathbf{q}^r_{i,l_r}\right] \ \ \in\ \ \mathbb{R}^{d \times l_r} \ ,
\end{eqnarray}
which correspond to our domain-specific feature extractors  $\mathbf{Q}_r(\mathbf{x}^r_i) \triangleq [q_r^1(\mathbf{x}^r_i) \ldots q_r^{l_r}(\mathbf{x}^r_i)]$ where $q_r^{t}(\mathbf{x}^r_i) = \mathbf{q}^r_{i, t} \in \mathbb{R}^d$, as defined previously in our transferable representation (Section~\ref{representation}).\\

\noindent{\bf Feature Scoring.} The concatenated features $\mathbf{Q}_r^{(i)}$ is then fed to the self-attention component $\mathbb{ATT}_r$ to generate a vector of importance scores for the output components, i.e. $\mathbf{a}_r^{(i)} \triangleq \mathbb{ATT}_r(\mathbf{Q}_r^{(i)}) \in \mathbb{R}^d$. For more details on how to construct this component, see ~\cite{chorowski2015attention,hermann2015teaching,ba2014multiple} and~\cite{lin2017structured}. The result corresponds to the feature scoring functions\footnote{We use the notation $[\mathbf{a}]_t$ to denote the $t$-th component of vector $\mathbf{a}$.} $\mathbf{A}_r(\mathbf{x}^r_i) \triangleq [a_r^{(1)}(\mathbf{x}^r_i)\ldots a_r^{(d)}(\mathbf{x}^r_i)]$ where $a_r^{(t)}(\mathbf{x}^r_i) = [\mathbf{a}_r^{(i)}]_t \in \mathbb{R}$.\\

\noindent {\bf Feature Aggregation.} The extracted features $\mathbf{Q}_r^{(i)}$ are combined using the above feature scoring functions, which yields $\mathbf{Q}_r^{(i)^\top}\mathbf{a}_r^{(i)}$. The combined features are subsequently passed through a linear layer with densely connected hidden units ($\mathbb{DENSE}_r$),
\begin{eqnarray}
\mathbf{g}^{(i)}_r &\triangleq& \mathbb{DENSE}_r\left(\mathbf{Q}_r^{(i)^\top}\mathbf{a}_r^{(i)}; \mathbf{w}_r\right) \ ,
\end{eqnarray}
where $\mathbf{g}^{(i)}_r \in \mathbb{R}^c$ with $c$ denotes the number of class labels and $\mathbf{w}_r$ denotes the parametric weights of the dense layers. Then, the output of the dense layer is transformed into a probability distribution over class labels via the following softmax activation functions parameterized with softmax temperatures $\tau_r$:
\begin{eqnarray}
\hspace{-2mm}\mathcal{T}\left(y = j|\mathbf{x}^r_i\right) \triangleq \mathrm{exp}\left(\left[\mathbf{g}^{(i)}_r\right]_j/\tau_r\right) \Bigg/ \sum_{\kappa =1}^c \mathrm{exp}\left(\left[\mathbf{g}^{(i)}_r\right]_{\kappa}/\tau_r\right)\nonumber
\hspace{-10mm}
\end{eqnarray}
The entire process corresponds to the feature aggregation function $\mathbf{O}_r(\mathbf{Q}_r(\mathbf{x}^r)^\top\mathbf{A}_r(\mathbf{x}^r))$ parameterized by $\{\mathbf{w}_r, \tau_r\}$.

\subsection{Knowledge Infusion for Poor Model}\label{sec:algorithm}
To infuse the above knowledge extracted from the \emph{rich} model to the \emph{poor} model, we adopt the same transferable representation for the \emph{poor} model as follows:
\begin{eqnarray*}
\mathcal{S}\left(y\ |\ \mathbf{x}^p\right) &\triangleq& \mathrm{Pr}\Big(y \ |\  \mathbf{Q}_p^\top\left(\mathbf{x}^p\right)\mathbf{A}_p\left(\mathbf{x}^p\right); \mathbf{O}_p\Big)
\end{eqnarray*}
where $\mathbf{Q}_p$, $\mathbf{A}_p(\mathbf{x}^p) \triangleq [a_p^{(1)}(\mathbf{x}^p; \boldsymbol{\omega}_p^{(1)})\ldots a_p^{(d)}(\mathbf{x}^p; \boldsymbol{\omega}_p^{(d)})]\in  \mathbb{R}^d$ and $\mathbf{O}_p$ are the \emph{poor} model's domain-specific feature extractors, feature scoring functions and feature aggregation functions, which are similar in format to those of the \emph{rich} model. Infusing knowledge from the \emph{rich} model to the \emph{poor} model can then be boiled down to matching these components between them. This process can be decomposed into two steps:\\

{\noindent\bf Behavior Infusion.} As mentioned above, each scoring function $\mathbf{a}_p^{(i)}\left(\mathbf{x}^p; \boldsymbol{\omega}_p^{(i)}\right)$ is defined by a weight vector $\boldsymbol{\omega}_p^{(i)}$. The collection of these weight vectors thus defines the \emph{poor model}'s \emph{learning behaviors} (i.e., its feature scoring mechanism).\\

\noindent Given the input components $\{(\mathbf{x}^p_t, \mathbf{x}^r_t)\}_{t=1}^k$ of the subjects included in the paired dataset $\mathcal{H}_o$ and the \emph{rich model}'s scoring outputs $\{a_r^{(i)}(\mathbf{x}_t^r)\}_{t=1}^k$ at those subjects, we can construct an auxiliary dataset $\mathcal{B}_i \triangleq \{(\mathbf{x}^p_t, a_r^{(i)}(\mathbf{x}^r_t))\}_{t=1}^k$ to learn the corresponding behavior $\boldsymbol{\omega}_p^{(i)}$ of the \emph{poor model} so that its scoring mechanism is similar to that of the \emph{rich model}. That is, we want to learn a mapping from a \emph{poor} data point $\mathbf{x}^p$ to the important score assigned to its $i^{\mathrm{th}}$ latent feature by the \emph{rich model}. Formally, this can be cast as the optimization task given by Eq.~\ref{eq:3}:\vspace{-1.4em}

\begin{align}\label{eq:3}
\begin{array}{lll}
\hspace{-1.8mm}\displaystyle \minimize_{\boldsymbol{\omega}_p^{(i)}}  \displaystyle\mathcal{L}_i\left(\bm{\omega}_p^{(i)}\right) \hspace{-0mm}&\triangleq&\hspace{-2mm} \displaystyle\frac{1}{2}\sum_{t=1}^k\left(a_p^{(i)}\left(\mathbf{x}^p_t; \boldsymbol{\omega}_p^{(i)}\right) - a_r^{(i)}\left(\mathbf{x}^r_t\right)\right)^2\\
\hspace{-0mm}&+&\hspace{-2mm} \lambda\|\boldsymbol{\omega}_p^{(i)}\|^2_2\ .
\end{array}
\end{align}
\vspace{-0.8em}

\noindent For example, if we parameterize $a_p^{(i)}\left(\mathbf{x}^p_t; \boldsymbol{\omega}_p^{(i)}\right) = \boldsymbol{\omega}_p^{{(i)}^\top}\mathbf{x}^p_t$ and choose $\lambda = 0$, then Eq.~\ref{eq:3} reduces to a linear regression task, which can be solved analytically. Alternatively, by choosing $\lambda = 1$, Eq.~\ref{eq:3} reduces to a maximum a posterior (MAP) inference task with normal prior imposed on $\boldsymbol{\omega}_p^{(i)}$, which is also analytically solvable.\\

\noindent Incorporating more sophisticated, non-linear parameterization for $a_p^{(i)}\left(\mathbf{x}^p_t; \boldsymbol{\omega}_p^{(i)}\right)$ (e.g., deep neural network with varying structures) is also possible but Eq.~\ref{eq:3} can only be optimized approximately via numerical methods (see Section~\ref{representation}). Eq.~\eqref{eq:3} can be solved via standard gradient descent. The complexity of deriving the solution thus depends on the number of iteration $\tau$ and the cost of computing the gradient of $\boldsymbol{\omega}_p^{(i)}$ which depends on the parameterization of $a_p^{(i)}$ but is usually $O(w)$ where $w = \max_i |\boldsymbol{\omega}_p^{(i)}|$. As such, the cost of computing the gradient of the objective function with respect to a particular $i$ is O(kw). As there are $\tau$ iterations, the cost of solving for the optimal $\boldsymbol{\omega}_p^{(i)}$ is $O(\tau k w)$. Lastly, since we are doing this for $d$ values of $i$, the total complexity would be $O(\tau k w d)$.\\

{\noindent\bf Target Infusion.} Given the \emph{poor model}'s learned behaviors $\{\boldsymbol{\omega}_p^{(i)}\}_{i=1}^d$ (which were fitted to those of the \emph{rich model} via solving Eq.~\ref{eq:3}), we now want to optimize the \emph{poor model}'s feature aggregation $\mathbf{O}_p$ and feature extraction $\mathbf{Q}_p$ components so that its predictions will (a) fit those of the \emph{rich model} on paired data $\mathcal{H}_o$; and also (b) fit the ground truth $\{y^p_t\}_{t=1}^m$ provided by the poor data $\mathcal{H}_p$. 
Formally, this can be achieved by solving the following optimization task:\vspace{-1mm}
\begin{align}\label{eq:4}
\begin{array}{lll}
\hspace{-3.5mm}\displaystyle \min_{\mathbf{O}_p, \mathbf{Q}_p} \displaystyle\mathcal{L}_p &\hspace{-3mm}\triangleq&\hspace{-3mm} \displaystyle\frac{1}{k}\sum_{t=1}^k\sum_{y=1}^c \left(\mathcal{T}\left(y_t|\mathbf{x}^r_t; \mathbf{O}_r, \mathbf{Q}_r\right) - \mathcal{S}\left(y|\mathbf{x}^p_t; \mathbf{O}_p, \mathbf{Q}_p\right)\right)^2 \\
&\hspace{-3mm}+&\hspace{-3mm}\displaystyle \frac{1}{m}\sum_{t=1}^m \left(1 - \mathcal{S}\left(y^p_t|\mathbf{x}^p_t; \mathbf{O}_p, \mathbf{Q}_p\right)\right)^2
\end{array}
\end{align}
To understand the above, note that the first term tries to fit \emph{poor model} $\mathcal{S}$ to \emph{rich model} $\mathcal{T}$ in the context of the paired dataset $\mathcal{H}_o \triangleq \{(\mathbf{x}_t^p,\mathbf{x}_t^r,y_t)\}_{t=1}^k$ while the second term tries to adjust the \emph{poor model}'s fitted behavior and target in a local context of its \emph{poor data} $\mathcal{H}_p \triangleq \{(\mathbf{x}^p_t, y^p_t)\}_{t=1}^m$. This allows the second term to act as a filter that downplays distilled patterns which are irrelevant in the poor data context. Again, Eq.~\ref{eq:4} can be solved depending on how we parameterize the aforementioned components $(\mathbf{O}_p, \mathbf{Q}_p)$.\\

\noindent For example, $\mathbf{O}_p$ can be set as a linear feed-forward layer with densely connected hidden units, which are activated by a softmax function. Again, Eq.~\eqref{eq:4} could be solved via standard gradient descent. The cost of computing the gradient would depend linearly on the total no. $n_p$ of neurons in the parameterization of $\mathbf{O}_p$ and $\mathbf{Q}_p$ for the poor model. In particular, the gradient computation complexity for one iteration is $O(n_p(mpc + kc^2))$. For $\tau$ iteration, the total cost would be $O(\tau n_p(mpc + kc^2))$.\\

\noindent Both steps of {\bf behavior infusion} and {\bf target infusion} are succinctly summarized in Algorithm~\ref{alg:alg} below.

\begin{algorithm}[ht]
\caption{\mname($\mathcal{H}_p$, $\mathcal{T}(y|\mathbf{x}^r)$, $\mathcal{H}_o$)}
\label{alg:alg}
\begin{algorithmic}[1]
\STATE Input: \emph{rich model} $\mathcal{T}(y|\mathbf{x}^r)$, \emph{poor data} $\mathcal{H}_p$ and paired data $\mathcal{H}_o$
\STATE Infuse \emph{rich model}'s behavior via $\mathcal{H}_o$
\STATE $i \leftarrow 1$
\WHILE {$i \leq d$}
\STATE $\bm{\omega}_p^{(i)}\leftarrow \argmin\mathcal{L}_i(\bm{\omega}_p^{(i)})$ via ~\eqref{eq:3}; 
\STATE $a_p^{(i)}(\mathbf{x}^p) \leftarrow a_p^{(i)}(\mathbf{x}^p; \bm{\omega}_p^{(i)})$
\STATE $i \leftarrow i + 1$
\ENDWHILE
\STATE $\mathbf{A}_p \ \leftarrow\   [a_p^{(1)}(\mathbf{x}^p;\bm{\omega}_p^{(1)}) \ldots a_p^{(d)}(\mathbf{x}^p;\bm{\omega}_p^{(d)})]$
\STATE Infuse \emph{rich model}'s target via $(\mathcal{H}_o, \mathcal{H}_p)$ and $\mathbf{A}_p$
\STATE $(\mathbf{Q}_p,\mathbf{O}_p) \  \leftarrow\  \argmin \mathcal{L}_p$ via~\eqref{eq:4}
\STATE Output: \emph{poor model} $\mathcal{S}(y|\mathbf{x}^p) \leftarrow (\mathbf{A}_p, \mathbf{Q}_p,\mathbf{O}_p)$ 
\end{algorithmic}
\end{algorithm}

\section{Theoretical Analysis}
\label{theory}
In this section, we provide theoretical analysis for \tname. Our goal is to show that under certain practical assumptions and with respect to a random instance $\mathbf{x} = (\mathbf{x}^p,\mathbf{x}^r) \sim \mathcal{P}(\mathbf{x})$ drawn from an arbitrary data distribution $\mathcal{P}$, the prediction $y^p \triangleq \argmax \mathcal{S}(y | \mathbf{x}^p)$ of the resulting \emph{poor model} will agree with that of the \emph{rich model}, $y^r \triangleq \argmax \mathcal{T}(y | \mathbf{x}^r)$, with high probability, thus demonstrating the accuracy of our knowledge infusion algorithm in Section~\ref{sec:algorithm}.\\

\noindent {\bf High-Level Ideas.} To achieve this, our strategy is to first bound the \emph{expected} target fitting loss (see Definition $2$) on a random instance $\mathbf{x} \triangleq (\mathbf{x}^p,\mathbf{x}^r) \sim \mathcal{P}(\mathbf{x})$ of the \emph{poor model} with respect to its optimized scoring function $\mathbf{A}_p$, feature extraction $\mathbf{Q}_p$ and feature aggregation $\mathbf{O}_p$ components via solving Eq.~\ref{eq:3} and Eq.~\ref{eq:4} in Section~\ref{sec:algorithm} (see Lemma $1$).\\

\noindent We can then characterize the sufficient condition on the target fitting loss (see Definition $1$) with respect to a particular instance $\mathbf{x} \triangleq (\mathbf{x}^p,\mathbf{x}^r)$ for the \emph{poor model} to agree with the \emph{rich model} on their predictions of $\mathbf{x}^p$ and $\mathbf{x}^r$, respectively (see Lemma $2$). The probability that this sufficient condition happens can then be bounded in terms of the bound on the \emph{expected} target fitting loss in Lemma $1$ (see Theorem $1$), which in turn characterizes how likely the \emph{poor model} will agree with the \emph{rich model} on the prediction of a random data instance. To proceed, we put forward the following assumptions and definitions:\\

\noindent {\bf Definition $1$.} Let $\boldsymbol{\theta}_p \triangleq \{\mathbf{O}_p,\mathbf{Q}_p,\mathbf{A}_p\}$ denote an arbitrary parameterization of the \emph{poor model}. The \emph{particular} target fitting loss of the \emph{poor model} with respect to a data instance $\mathbf{x} \triangleq (\mathbf{x}^p,\mathbf{x}^r)$ is
\begin{eqnarray}
\hspace{-50mm}\widehat{\mathcal{L}}_{\mathbf{x}}\left(\boldsymbol{\theta}_p\right) &\triangleq& \sum_{y=1}^c \Big(\mathcal{T}\left(y|\mathbf{x}^r\right) - \mathcal{S}\left(y|\mathbf{x}^p\right)\Big)^2 \nonumber\\
&+& \frac{1}{m}\sum_{t=1}^m \Big(1 - \mathcal{S}\left(y^p_t|\mathbf{x}^p_t\right)\Big)^2 \ , \label{eq:5}
\end{eqnarray}
where $c$ denotes the number of classes, $\mathcal{T}(y|\mathbf{x}^r)$ and $\mathcal{S}(y|\mathbf{x}^p)$ denotes the probability scores assigned to candidate class $y$ by the \emph{rich} and \emph{poor} models, respectively.\\ 

\noindent {\bf Definition $2$.} Let $\boldsymbol{\theta}_p$ be defined as in Definition $1$. The \emph{expected} target fitting loss of the \emph{poor model} with respect to the parameterization $\boldsymbol{\theta}_p$ is defined below,
\begin{eqnarray}
\mathcal{L}\left(\boldsymbol{\theta}_p\right) &\triangleq& \mathbb{E}_{\mathbf{x} \sim \mathcal{P}(\mathbf{x})}\left[\widehat{\mathcal{L}}_{\mathbf{x}}\left(\boldsymbol{\theta}_p\right)\right] \ , \label{eq:6}
\end{eqnarray}
where the expectation is over the unknown data distribution $\mathcal{P}(\mathbf{x})$.\\

\noindent {\bf Definition $3$.} Let $\mathbf{x} = (\mathbf{x}^p,\mathbf{x}^r)$ and $y(\mathbf{x}) \triangleq \argmax_{y=1}^c \mathcal{T}(y|\mathbf{x}^r)$. The robustness constant of the \emph{rich model} is defined below,
\begin{eqnarray}
\hspace{-8mm}\phi &\triangleq& \frac{1}{2}\min_{(\mathbf{x}^p,\mathbf{x}^r)}\left(\mathcal{T}\left(y(\mathbf{x}) \ |\ \mathbf{x}^r\right) - \max_{y \ne y(\mathbf{x})}\mathcal{T}\left(y \ |\ \mathbf{x}^r\right)\right)
\ , \label{eq:7}
\end{eqnarray}
That is, if the probability scores of the model are being perturbed additively within $\phi$, its prediction outcome will not change.\\ 

\noindent {\bf Assumption $1$.} The paired data points $\mathbf{x}_i = (\mathbf{x}_i^p,\mathbf{x}^r_i)$ of $\mathcal{H}_o$ are assumed to be distributed independently and identically from $\mathcal{P}(\mathbf{x})$.\\

\noindent {\bf Assumption $2$.} The hard-label predictions $y^p \triangleq \argmax \mathcal{S}(y|\mathbf{x}^p)$ and $y^r \triangleq \argmax \mathcal{T}(y|\mathbf{x}^r)$ of the \emph{poor} and \emph{rich} models are unique.\\

\noindent \textit{Given the above, we are now ready to state our first result}:\\

\noindent {\bf Lemma $1$.} Let $\boldsymbol{\theta}_p^\ast$ and $\widehat{\boldsymbol{\theta}}_p$ denote the optimal parameterization of the \emph{poor model} that yields the minimum \emph{expected} target fitting loss (see Definition $2$) and the optimal solution found by minimizing the objective functions in Eq.~\ref{eq:3} and Eq.~\ref{eq:4}, respectively. Let $\alpha \triangleq \mathcal{L}(\boldsymbol{\theta}_p^\ast)$, $\delta \in (0, 1)$ and $c$ denote the number of classes in our predictive task. If $k \triangleq |\mathcal{H}_o| \geq ((c + 1)^2/(2\epsilon^2))\log(2/\delta)$ then, 
\begin{eqnarray}
\mathrm{Pr}\left(\mathcal{L}\left(\widehat{\boldsymbol{\theta}}_p\right) \ \leq\  \alpha \ +\ 2\epsilon\right) &\geq& 1 - \delta \ . \label{eq:8}
\end{eqnarray}

\noindent {\bf Proof.} We first note that by definition in Eq.~\eqref{eq:5}, for all $\mathbf{x}$, $\widehat{\mathcal{L}}_{\mathbf{x}}(\boldsymbol{\theta}) \leq c + 1$. Then, let us define the \emph{empirical} target fitting loss as
\begin{eqnarray}
\widehat{\mathcal{L}}\left(\boldsymbol{\theta}\right) &\triangleq& \frac{1}{k}\sum_{i=1}^k \widehat{\mathcal{L}}_{\mathbf{x}^{(i)}}\left(\boldsymbol{\theta}\right) \ , \label{eq:8c}
\end{eqnarray}
where $\{\widehat{\mathcal{L}}_{\mathbf{x}^{(i)}}\left(\boldsymbol{\theta}\right)\}_{i=1}^k$ can be treated as identically and independently distributed random variables in $(0, c + 1)$. Then, by Definition $2$, it also follows that $\mathcal{L}(\boldsymbol{\theta}) = \mathbb{E}[\widehat{\mathcal{L}}(\boldsymbol{\theta})]$. Thus, by Hoeffding inequality:
\begin{eqnarray}
\hspace{-7mm}\mathrm{Pr}\left(\left|\mathcal{L}(\boldsymbol{\theta}) - \widehat{\mathcal{L}}(\boldsymbol{\theta})\right| \leq \epsilon\right) &\geq& 1 - 2\mathrm{exp}\left(-\frac{2k\epsilon^2}{(c + 1)^2}\right) \ . \label{eq:8d}
\end{eqnarray}
Then, for an arbitrary $\delta \in (0, 1)$, setting $\delta \leq \mathrm{exp}(-2k\epsilon^2/(c + 1)^2)$ and solving for $k$ yields $k \geq ((c+1)^2/(2\epsilon^2))\log(2/\delta)$. Thus, for $k \geq ((c+1)^2/(2\epsilon^2))\log(2/\delta)$, with probability at least $1 - \delta$, $|\mathcal{L}(\boldsymbol{\theta}) - \widehat{\mathcal{L}}(\boldsymbol{\theta})| \leq \epsilon$ holds simultaneously for all $\boldsymbol{\theta}$. When that happens with probability at least $1 - \delta$, we have:
\begin{eqnarray}
\hspace{-9mm}\mathcal{L}\left(\widehat{\boldsymbol{\theta}}_p\right) &\leq& \widehat{\mathcal{L}}\left(\widehat{\boldsymbol{\theta}}_p\right) + \epsilon\nonumber\\
&\leq& \widehat{\mathcal{L}}\left(\boldsymbol{\theta}^\ast_p\right) + \epsilon \ \leq\ \mathcal{L}\left(\boldsymbol{\theta}_p^\ast\right) + 2\epsilon \ \ =\ \ \alpha + 2\epsilon \ .
\end{eqnarray}
That is, $\mathrm{Pr}\left(\widehat{\mathcal{L}}\left(\widehat{\boldsymbol{\theta}}_p\right) \leq \alpha + 2\epsilon\right) \ \geq\  1 - \delta$, which completes our proof for Lemma $1$. Note that the above 2nd inequality follows from the  definition of $\widehat{\bm{\theta}}_p \triangleq \argmin_{\bm{\theta}} \widehat{\mathcal{L}}(\bm{\theta}_p)$, which implies $\widehat{\mathcal{L}}(\widehat{\bm{\theta}}_p) \leq \widehat{\mathcal{L}}(\bm{\theta}^\ast_p)$.\vspace{1mm}

\noindent This result implies the \emph{expected} target fitting loss $\mathcal{L}(\widehat{\theta}_p)$ incurred by our knowledge infusion algorithm in Section~\ref{sec:algorithm} can be made arbitrarily close (with high confidence) to the optimal \emph{expected} target fitting loss $\alpha \triangleq \mathcal{L}(\boldsymbol{\theta}_p^\ast)$ with a sufficiently large paired dataset $\mathcal{H}_o$.\\

\noindent {\bf Lemma $2$.} Let $\mathbf{x} = (\mathbf{x}^p,\mathbf{x}^r)$ and $\widehat{\boldsymbol{\theta}}_p$ as defined in Lemma $1$. If the corresponding \emph{particular} target fitting loss (see Definition $1$) $\widehat{\mathcal{L}}_{\mathbf{x}}(\widehat{\boldsymbol{\theta}}_p) \leq \phi^2$, then both \emph{poor} and \emph{rich} models agree on their predictions for $\mathbf{x}^p$ and $\mathbf{x}^r$, respectively. That is, $y^p \triangleq \max_y \mathcal{S}(y|\mathbf{x}^p)$ and $y^r \triangleq \max_y \mathcal{T}(y|\mathbf{x}^r)$ are the same.\\

\noindent {\bf Proof.} Let $y^p$ and $y^r$ be defined as in the statement of Lemma $2$. We have:
\begin{eqnarray}
\hspace{-8mm}\mathcal{S}\left(y^p \ |\ \mathbf{x}^p\right) \hspace{-2mm}&\geq&\hspace{-2mm} \mathcal{S}\left(y^r \ |\ \mathbf{x}^p\right) \ \ \geq\ \ \mathcal{T}\left(y^r \ |\ \mathbf{x}^r\right) - \phi \nonumber\\
\hspace{-2mm}&\geq&\hspace{-2mm} \mathcal{T}\left(y^p \ |\ \mathbf{x}^r\right) + 2\phi - \phi \nonumber\\
\hspace{-2mm}&\geq&\hspace{-2mm} \mathcal{S}\left(y^p \ |\ \mathbf{x}^p\right) + 2\phi - 2\phi \ \ =\ \ \mathcal{S}\left(y^p \ |\ \mathbf{x}^p\right) \ .\label{eq:8a}
\end{eqnarray}
To understand Eq.~\eqref{eq:8a}, note that the first inequality follows from the definition of $y^p$. The second inequality follows from the fact that $\widehat{\mathcal{L}}_{\mathbf{x}}(\widehat{\boldsymbol{\theta}}_p) \leq \phi^2$, which implies $\forall y \ (\mathcal{S}(y|\mathbf{x}^p) -\mathcal{T}(y|\mathbf{x}^r))^2 \leq \phi^2$ and hence, $|\mathcal{S}(y^r|\mathbf{x}^p) - \mathcal{T}(y^r|\mathbf{x}^r)| \leq \phi$ or $\mathcal{S}(y^r|\mathbf{x}^p) \geq \mathcal{T}(y^r|\mathbf{x}^r) - \phi$. The third inequality follows from the definitions of $\phi$ (see Definition $3$) and $y^r$. Finally, the last inequality follows from the definition of $y^p$ and that $\widehat{\mathcal{L}}_{\mathbf{x}}(\widehat{\boldsymbol{\theta}}_p) \leq \phi^2$, which also implies $\mathcal{T}(y^p|\mathbf{x}^r) \geq \mathcal{S}(y^p|\mathbf{x}^p) - \phi$.\\ 

\noindent Eq.~\eqref{eq:8a} thus implies $\mathcal{S}\left(y^p | \mathbf{x}^p\right) \geq \mathcal{S}\left(y^r | \mathbf{x}^p\right) \geq \mathcal{S}\left(y^p | \mathbf{x}^p\right)$ and hence, $\mathcal{S}\left(y^p | \mathbf{x}^p\right) = \mathcal{S}\left(y^r | \mathbf{y}^p\right)$. Since the hard-label prediction is unique (see Assumption $3$), this means $y^r = y^p$ and hence, by definitions of $y^r$ and $y^p$, the \emph{poor} and \emph{rich} models yield the same prediction. This completes our proof for Lemma $2$.

\noindent Intuitively, Lemma $2$ specifies the sufficient condition under which the \emph{poor model} will yield the same hard-label prediction on a particular data instance $\mathbf{x}$ as the \emph{rich model}. Thus, if we know how likely this sufficient condition will happen, we will also know how likely the \emph{poor model} will imitate the \emph{rich model} successfully on a random data instance. This intuition is the key result of our theoretical analysis and is formalized below:\\

\noindent {\bf Theorem $1$.} Let $\delta \in (0, 1)$ and $\mathbf{x} = (\mathbf{x}^p, \mathbf{x}^r)$ denote a random instance drawn from $\mathcal{P}(\mathbf{x})$. Let $k \triangleq |\mathcal{H}_o|$ denote the size of the paired dataset $\mathcal{H}_o$, which were used to fit the learning behaviors of the \emph{poor model} to that of the \emph{rich model}, and $\mathcal{E}$ denotes the event that both models agree on their predictions of $\mathbf{x}$. If $k \geq ((c+1)^2/(2\epsilon^2))\log(2/\delta)$, then with probability at least $1 - \delta$,
\begin{eqnarray}
\mathrm{Pr}\left(\mathcal{E}\right) &\geq& 1 - \frac{1}{\phi^2}\left(\alpha + 2\epsilon\right) \ .\label{eq:9}
\end{eqnarray}

\noindent {\bf Proof.} Since $\widehat{\mathcal{L}}_{\mathbf{x}}(\widehat{\boldsymbol{\theta}}_p) \leq \phi^2$ implies $\mathcal{E}$, it follows that 
\begin{eqnarray}
\mathrm{Pr}\left(\mathcal{E}\right) &\geq& \mathrm{Pr}\left(\widehat{\mathcal{L}}_{\mathbf{x}}\left(\widehat{\boldsymbol{\theta}}_p\right) \leq \phi^2\right) \ . \label{eq:10}
\end{eqnarray} 
Then, by Markov inequality, we have
\begin{eqnarray}
\hspace{-8mm}\mathrm{Pr}\left(\widehat{\mathcal{L}}_{\mathbf{x}}\left(\widehat{\boldsymbol{\theta}}_p\right) > \phi^2\right) \hspace{-3mm}&\leq&\hspace{-3mm} \phi^{-2}\mathbb{E}\left[\widehat{\mathcal{L}}_{\mathbf{x}}\left(\widehat{\boldsymbol{\theta}}_p\right)\right] =  \phi^{-2}\mathcal{L}\left(\widehat{\theta}_p\right) \ . \label{eq:11}
\end{eqnarray}
Subtracting both sides of Eq.~\eqref{eq:11} from a unit probability yields
\begin{eqnarray}
\mathrm{Pr}\left(\widehat{\mathcal{L}}_{\mathbf{x}}\left(\widehat{\boldsymbol{\theta}}_p\right) \leq \phi^2\right) &\geq& 1 \ -\  \phi^{-2}\mathcal{L}\left(\widehat{\theta}_p\right) \ , \label{eq:12} 
\end{eqnarray}
where the last equality follows because $\mathbb{E}[\widehat{\mathcal{L}}_{\mathbf{x}}(\widehat{\boldsymbol{\theta}}_p)] = \mathcal{L}(\widehat{\boldsymbol{\theta}}_p)$, which follows immediately from Definitions $1$-$2$ and Assumption $1$. Thus, plugging Eq.~\eqref{eq:12} into Eq.~\eqref{eq:10} yields
\begin{eqnarray}
\mathrm{Pr}\left(\mathcal{E}\right) &\geq& 1 \ -\  \phi^{-2}\mathcal{L}\left(\widehat{\theta}_p\right) \ . \label{eq:13} 
\end{eqnarray}
Applying Lemma $2$, we know that with probability $1 - \delta$, $\mathcal{L}\left(\widehat{\theta}_p\right) \leq \alpha + 2\epsilon$. Thus, plugging this into Eq.~\eqref{eq:13} yields
\begin{eqnarray}
\mathrm{Pr}\left(\mathcal{E}\right) &\geq& 1 \ -\  \phi^{-2}\left(\alpha + 2\epsilon\right) \ . \label{eq:14} 
\end{eqnarray}
That is, by union bound, with probability at least $1 - \delta - \phi^{-2}(\alpha + 2\epsilon)$, the \emph{poor model} yields the same prediction as that of the \emph{rich model}. This completes our proof for Theorem $1$.

\noindent This immediately implies $\mathcal{E}$ will happen with probability at least $1 - \delta - (1/\phi^2)(\alpha + 2\epsilon)$. The chance for the \emph{poor model} to yield the same prediction as the \emph{rich model} on an arbitrary instance (i.e., knowledge infusion succeeds) is therefore at least $1 - \delta - (1/\phi^2)(\alpha + 2\epsilon)$.

\section{Experiments}
\label{experiment}

\subsection{Experimental Settings}

\textbf{Datasets.} We use the following datasets in our evaluation.\\

\noindent {\bf A. MIMIC-III Critical Care Database (MIMIC-III) \footnote{\url{https://mimic.physionet.org/}}} is collected from more than $58,000$ ICU patients at the Beth Israel Deaconess Medical Center (BIDMC) from June 2001 to October 2012 ~\cite{johnson2016mimic}.  We collect a subset of $9,488$ patients who has one of the following (most frequent) diseases in their main diagnosis: (1) acute myocardial infarction, (2) chronic ischemic heart disease, (3) heart failure, (4) intracerebral hemorrhage, (5) specified procedures complications, (6) lung diseases, (7) endocardium diseases, and (8) septicaemia. The task is disease diagnosis classification (i.e., predicting which of 8 diseases the patient has) based on features collected from 6 data channels: vital sign time series including Heart Rate (HR), Respiratory Rate (RR), Blood Pressure mean (BPm), Blood Pressure systolic (BPs), Blood Pressure diastolic (BPd) and Blood Oxygen Saturation (SpO2). We randomly divided the data into training ($80\%$), validation ($10\%$) and testing ($10\%$) sets. \\

\noindent {\bf B. PTB Diagnostic ECG Database (PTBDB) \footnote{\url{https://physionet.org/physiobank/database/ptbdb/}}} is a 15-channel 1000 Hz ECG time series including 12 conventional leads and 3 Frank leads ~\cite{goldberger2000physiobank,bousseljot1995nutzung} collected from both healthy controls and cases of heart diseases, which amounts to a total number of 549 records.
The given task is to classify ECG to one of the following categories: (1) myocardial infarction, (2) healthy control, (3) heart failure, (4) bundle branch block, (5) dysrhythmia, and (6) hypertrophy.
We down-sampled the data to 200 Hz and pre-processed it following the "frame-by-frame" method~\cite{reiss2012creating} with sliding windows of 10-second duration and 5-second stepping between adjacent windows.
\\

\noindent {\bf C. NEDC TUH EEG Artifact Corpus (EEG)}
\footnote{\url{https://www.isip.piconepress.com/projects/tuh_eeg/html/overview.shtml}} is a 22-channel 500 Hz sensor time series collected from over 30,000 EEGs spanning the years from 2002 to present ~\cite{obeid2016temple}. The task is to classify 5 types of EEG events including (1) eye movements (EYEM), (2) chewing (CHEW), (3) shivering (SHIV), (4) electrode pop, electrode static, and lead artifacts (ELPP), and (5) muscle artifacts (MUSC). We randomly divided the data into training ($80\%$), validation ($10\%$) and testing ($10\%$) sets by records.\\

\noindent The statistics of the above datasets, as well as the architectures of the \emph{rich} and \emph{poor} models on each dataset are summarized in the tables below.

\begin{table}[ht]
\centering
\caption{Data statistics. }
\label{table:data}
\resizebox{0.9\columnwidth}{!}{
\begin{tabular}{lrrr}
\toprule
                            & MIMIC-III     & PTBDB         & EEG        \\
\midrule
\hspace{-2mm}\# subjects                 & 9,488         & 549           & 213        \\
\hspace{-2mm}\# classes                  & 8             & 6             & 5          \\
\hspace{-2mm}\# features               & 6             & 15            & 22         \\
\hspace{-2mm}Average length              & 48            & 108,596       & 13,007     \\
\hspace{-2mm}Sample frequency            & 1 per hour    & 1,000 Hz      & 500 Hz     \\
\bottomrule
\end{tabular}
}
\end{table}

\begin{table}[!ht]
\centering
\caption{The architecture of \emph{rich} model in MIMIC-III, which includes a total of 51.6k parameters. }
\resizebox{\columnwidth}{!}{
\begin{tabular}{rlll}
\toprule
\multicolumn{1}{l}{Layer} & Type              & Hyper-parameters                       & Activation  \\
\midrule
1                         & Split             & n\_seg=6                               &             \\
2                         & Convolution1D     & n\_filter=64, kernel\_size=4, stride=1 & ReLU        \\
3                         & Convolution1D     & n\_filter=64, kernel\_size=4, stride=1 & ReLU        \\
4                         & AveragePooling1D  &                                        &             \\
5                         & LSTM              & hidden\_units=64                       & ReLU        \\
6                         & PositionAttention &                                        &             \\
7                         & Dense             & hidden\_units=n\_classes               & Linear      \\
8                         & Softmax           &                                        &            \\
\bottomrule
\end{tabular}
}
\end{table}

\begin{table}[!ht]
\centering
\caption{The architecture of the infused, \emph{poor} model used by \tname, Direct, KD and AT for knowledge infusion in MIMIC-III, which includes a total of 3.5k parameters. }
\resizebox{\columnwidth}{!}{
\begin{tabular}{rlll}
\toprule
\multicolumn{1}{l}{Layer} & Type              & Hyper-parameters                       & Activation  \\
\midrule
1                         & Split             & n\_seg=6                               &             \\
2                         & Convolution1D     & n\_filter=16, kernel\_size=4, stride=1 & ReLU        \\
3                         & Convolution1D     & n\_filter=16, kernel\_size=4, stride=1 & ReLU        \\
4                         & AveragePooling1D  &                                        &             \\
5                         & LSTM              & hidden\_units=16                       & ReLU        \\
6                         & PositionAttention &                                        &             \\
7                         & Dense             & hidden\_units=n\_classes               & Linear      \\
8                         & Softmax           &                                        &            \\
\bottomrule
\end{tabular}
}
\end{table}

\begin{table}[!ht]
\centering
\caption{The architecture of \emph{rich} model in PTBDB, which includes a total of 688.8k parameters. }
\label{tb:ptbdb_architecture}
\resizebox{\columnwidth}{!}{
\begin{tabular}{rlll}
\toprule
\multicolumn{1}{l}{Layer} & Type              & Hyper-parameters                       & Activation  \\
\midrule
1                         & Split             & n\_seg=10                               &             \\
2                         & Convolution1D     & n\_filter=128, kernel\_size=16, stride=2 & ReLU        \\
3                         & Convolution1D     & n\_filter=128, kernel\_size=16, stride=2 & ReLU        \\
4                         & Convolution1D     & n\_filter=128, kernel\_size=16, stride=2 & ReLU        \\
5                         & AveragePooling1D  &                                        &             \\
6                         & LSTM              & hidden\_units=128                       & ReLU        \\
7                         & PositionAttention &                                        &             \\
8                         & Dense             & hidden\_units=n\_classes               & Linear      \\
9                         & Softmax           &                                        &            \\
\bottomrule
\end{tabular}
}
\end{table}

\begin{table}[!ht]
\centering
\caption{The architecture of the infused, \emph{poor} model used by \tname, Direct, KD and AT for knowledge infusion in PTBDB, which includes a total 45.0k parameters. }
\label{tb:ptbdb_architecture_poor}
\resizebox{\columnwidth}{!}{
\begin{tabular}{rlll}
\toprule
\multicolumn{1}{l}{Layer} & Type              & Hyper-parameters                       & Activation  \\
\midrule
1                         & Split             & n\_seg=10                               &             \\
2                         & Convolution1D     & n\_filter=32, kernel\_size=16, stride=2 & ReLU        \\
3                         & Convolution1D     & n\_filter=32, kernel\_size=16, stride=2 & ReLU        \\
4                         & Convolution1D     & n\_filter=32, kernel\_size=16, stride=2 & ReLU        \\
5                         & AveragePooling1D  &                                        &             \\
6                         & LSTM              & hidden\_units=32                       & ReLU        \\
7                         & PositionAttention &                                        &             \\
8                         & Dense             & hidden\_units=n\_classes               & Linear      \\
9                         & Softmax           &                                        &            \\
\bottomrule
\end{tabular}
}
\end{table}

\begin{table}[b]
\centering
\caption{The architecture of \emph{rich} model in EEG, which includes a total of 417.4k parameters. }
\resizebox{\columnwidth}{!}{
\begin{tabular}{rlll}
\toprule
\multicolumn{1}{l}{Layer} & Type              & Hyper-parameters                       & Activation  \\
\midrule
1                         & Split             & n\_seg=5                               &             \\
2                         & Convolution1D     & n\_filter=128, kernel\_size=8, stride=2 & ReLU        \\
3                         & Convolution1D     & n\_filter=128, kernel\_size=8, stride=2 & ReLU        \\
4                         & Convolution1D     & n\_filter=128, kernel\_size=8, stride=2 & ReLU        \\
5                         & AveragePooling1D  &                                        &             \\
6                         & LSTM              & hidden\_units=128                       & ReLU        \\
7                         & PositionAttention &                                        &             \\
8                         & Dense             & hidden\_units=n\_classes               & Linear      \\
9                         & Softmax           &                                        &            \\
\bottomrule
\end{tabular}
}
\end{table}

\begin{table}[b]
\centering
\caption{The architecture of the infused, \emph{poor} model used by \tname, Direct, KD and AT for knowledge infusion in EEG, which includes a total of 51.6k parameters. }
\resizebox{\columnwidth}{!}{
\begin{tabular}{rlll}
\toprule
\multicolumn{1}{l}{Layer} & Type              & Hyper-parameters                       & Activation  \\
\midrule
1                         & Split             & n\_seg=5                               &             \\
2                         & Convolution1D     & n\_filter=32, kernel\_size=8, stride=2 & ReLU        \\
3                         & Convolution1D     & n\_filter=32, kernel\_size=8, stride=2 & ReLU        \\
4                         & Convolution1D     & n\_filter=32, kernel\_size=8, stride=2 & ReLU        \\
5                         & AveragePooling1D  &                                        &             \\
6                         & LSTM              & hidden\_units=32                       & ReLU        \\
7                         & PositionAttention &                                        &             \\
8                         & Dense             & hidden\_units=n\_classes               & Linear      \\
9                         & Softmax           &                                        &            \\
\bottomrule
\end{tabular}
}
\end{table}

\noindent\textbf{Baselines.}  We compare \mname against the following baselines:\\

\noindent \textbf{Direct}: In all experiments, we train a neural network model parameterized with \mname directly on the poor dataset without knowledge infusion from the \emph{rich model}. The resulting model can be used to produce a lower bound of predictive performance on each dataset.\\

\noindent \textbf{Knowledge Distilling (KD)}~\cite{hinton2015distilling}: KD transfers predictive power from teacher to student models via soft labels produced by the teacher model. In our experiments, all KD models have similar complexity as the infused model generated by \tname. The degree of label softness (i.e., the temperature parameter of soft-max activation function) in KD is set to 5. \\

\noindent \textbf{Attention Transfer (AT)}~\cite{Zagoruyko2017AT}: AT enhances shallow neural networks by leveraging attention mechanism ~\cite{bahdanau2014neural} to learn a similar attention behavior of a full-fledged deep neural network (DNN). In our experiments, we first train a DNN with attention component, which can be parameterized by \tname. The trained attention component of DNN is then transferred to that of a shallow neural networks in poor-data environment via activation-based attention transfer with $\mathrm{L}_2$-normalization.\\

\noindent \textbf{Heterogeneous Domain Adaptation (HDA)}~\cite{yao2019heterogeneous}: Maximize Mean Discrepancy (MMD) loss~\cite{GrettonBRSS06} has been successfully used in domain adaptation such as ~\cite{LongC0J15}. However, one drawback is that these works only consider homogeneous settings where the source and target domains have the same feature space, or use the same architecture of neural network. To mitigate this limitation, HDA~\citep{yao2019heterogeneous} proposed modification of soft MMD loss to handle with heterogeneity between source domain and target domain.  \\

\noindent{\bf  Performance Metrics.} The tested methods' prediction performance was compared based on their corresponding areas under the Precision-Recall (PR-AUC) and Receiver Operating Characteristic curves (ROC-AUC) as well as the accuracy and F1 score , which are often used in multi-class classification to evaluate the tested method's prediction quality. In particular, accuracy is measured by the ratio between the number of correctly classified instances and the total number of test instances. F1 score is the harmonic average of precision (the proportion of true positive cases among the predicted positive cases) and recall (the proportion of positive cases whose are correctly identified), with threshold $0.5$ to determine whether a predictive probability for being positive is large enough (larger than threshold) to actually assign a positive label to the case being considered or not.\\ 

\noindent Then, we use the average of F1 scores evaluated for each label (i.e., macro-F1 score) to summarize the averaged predictive performance of all tested methods across all classes. The ROC-AUC and PR-AUC scores are computed based on predicted probabilities and ground-truths directly. For ROC-AUC, it is the area under the curve produced by points of true positive rate (TPR) and false positive rate (FPR) at various threshold settings. Likewise, the PR-AUC score is the area under the curve produced by points of (precision, recall) at various threshold settings. In our experiments, we report the average PR-AUC and ROC-AUC since all three tasks are multi-class classification. \\

\noindent\textbf{Training Details} \noindent For each method, the reported results (mean performance and its empirical standard deviation) are averaged over 20 independent runs. For each run, we randomly split the entire dataset into training (80\%), validation (10\%) and test sets (10\%). All models are built using the training and validation sets and then, evaluated using test set. We use Adam optimizer ~\cite{adam} to train each model, with the default learning rate set to 0.001. The number of training epoches for each model is set as 200 and an early stopping criterion is invoked if the performance does not improve in 20 epoches. All models are implemented in Keras with Tensorflow backend and tested on a system equipped with 64GB RAM, 12 Intel Core i7-6850K 3.60GHz CPUs and Nvidia GeForce GTX 1080. For fair comparison, we use the same model architecture and hyper-parameter setting for Direct, KD, AT, HDA and \tname.  For rich dataset, we use the entire amount of dataset with the entire set of data features. For poor dataset, we vary the size of paired dataset and the number of features to analyze the effect of knowledge infusion in different data settings as shown in Section 4.3. The default maximum amount of paired data is set to 50\% of entire dataset, and the default number of data features used in the poor dataset is set to be half of the entire set of data features. In Section 4.2, to compare the tested methods' knowledge infusion performance under different data settings, we use the default settings for all models (including \tname and other baselines).

\subsection{Performance Comparison}

Results on MIMIC-III, PTBDB and EEG datasets are reported in Table ~\ref{tb:compare_mimic}, Table~\ref{tb:compare_ptbdb} and Table~\ref{tb:compare_eeg}, respectively. In this experiment, we set the size of paired dataset to 50\% of the size of the \emph{rich data}, and set the number of features used in poor-data environment to 3, 7, 11 for MIMIC-III, PTBDB and EEG, respectively. In all datasets, it can be observed that the infused model generated by \mname consistently achieves the best predictive performance among those of the tested methods, which demonstrates the advantage of our knowledge infusion framework over existing transfer methods such as KD and AT.\\

\noindent Notably, in terms of the macro-F1 scores, \tname improves over KD, AT, HDA and Direct by $5.60\%$, $23.95\%$, $31.84\%$ and $46.80\%$, respectively, on MIMIC-III dataset. The infused model generated by \tname also achieves $81.69\%$ performance of the \emph{rich} model on PTBDB in terms of the macro-F1 score (i.e., $0.299/0.366$, see Table~\ref{tb:compare_ptbdb}) while adopting an architecture that is $15.14$ times smaller than the rich model's (see Tables~\ref{tb:ptbdb_architecture} and~\ref{tb:ptbdb_architecture_poor}). We have also performed a significance test to validate the significance of our reported improvement of \mname over the baselines in Table~\ref{tb:p_value}. \\


\noindent Furthermore, it can also be observed that the performance variance of the infused model generated by \tname (as reflected in the reported standard deviation) is the lowest among all tested methods', which suggests that \tname's knowledge infusion is more robust. Our investigation in Section~\ref{sec:analysis} further shows that this is the result of \tname being able to perform both target and behavior infusion. This helps the infused model generated by \tname achieved better and more stable performance than those of KD, HDA and AT, which either match the prediction target or reasoning behavior of the \emph{rich} and \emph{poor} models (but not both). This consequently leads to their less robust performance with wide fluctuation in different data settings, as demonstrated next in Section~\ref{sec:analysis}.

\begin{table}[h!]
\caption{Performance comparison on MIMIC-III dataset.\vspace{-2mm}
}
\resizebox{\columnwidth}{!}{
\begin{tabular}{lllll}
\toprule
{} &          ROC-AUC &           PR-AUC &         Accuracy &         Macro-F1 \\
\midrule
Direct     &  0.622$\pm$0.062 &  0.208$\pm$0.044 &  0.821$\pm$0.012 &  0.141$\pm$0.057 \\
KD         &  0.686$\pm$0.043 &  0.257$\pm$0.029 &  0.833$\pm$0.012 &  0.196$\pm$0.049 \\
AT         &  0.645$\pm$0.064 &  0.225$\pm$0.044 &  0.826$\pm$0.013 &  0.167$\pm$0.057 \\ 
HDA         &  0.655$\pm$0.034 &  0.225$\pm$0.029 &  0.824$\pm$0.011 &  0.157$\pm$0.038 \\ 
\tname      &  {\bf 0.697$\pm$0.024} &  {\bf 0.266$\pm$0.023} &   {\bf 0.835$\pm$0.010} &   {\bf 0.207$\pm$0.030} \\
\hline
Rich Model &  0.759$\pm$0.014 &  0.341$\pm$0.024 &  0.852$\pm$0.007 &  0.295$\pm$0.027 \\
\bottomrule\vspace{-10mm}
\end{tabular}
}
\label{tb:compare_mimic}
\end{table}

\begin{table}[h!]
\caption{Performance comparison on PTBDB dataset.\vspace{-2mm}}
\resizebox{\columnwidth}{!}{
\begin{tabular}{lllll}
\toprule
{} &          ROC-AUC &           PR-AUC &         Accuracy &         Macro-F1 \\
\midrule
Direct     &  0.686$\pm$0.114 &  0.404$\pm$0.088 &  0.920$\pm$0.015 &  0.275$\pm$0.057 \\
KD         &  0.714$\pm$0.096 &  0.439$\pm$0.093 &  0.925$\pm$0.016 &  0.295$\pm$0.043 \\
AT         &  0.703$\pm$0.117 &  0.402$\pm$0.078 &  0.921$\pm$0.016 &  0.283$\pm$0.056 \\
HDA         &  0.685$\pm$0.113 &  0.430$\pm$0.080 &  0.924$\pm$0.011 &  {\bf 0.299$\pm$0.051} \\
\tname       &  {\bf 0.724$\pm$0.103} &   {\bf 0.441$\pm$0.080} &  {\bf 0.927$\pm$0.017} &  {\bf 0.299$\pm$0.052} \\
\hline
Rich Model &   0.732$\pm$0.110 &  0.483$\pm$0.101 &   0.930$\pm$0.017 &  0.366$\pm$0.071 \\
\bottomrule\vspace{-0mm}
\end{tabular}
}
\label{tb:compare_ptbdb}
\end{table}

\begin{table}[h!]
\caption{Performance comparison on EEG dataset.\vspace{-2mm}
}
\resizebox{\columnwidth}{!}{
\begin{tabular}{lllll}
\toprule
{} &          ROC-AUC &           PR-AUC &         Accuracy &         Macro-F1 \\
\midrule
Direct     &  0.797$\pm$0.064 &  0.506$\pm$0.083 &  0.888$\pm$0.015 &  0.425$\pm$0.078 \\
KD         &  0.772$\pm$0.083 &  0.512$\pm$0.082 &  0.888$\pm$0.021 &  0.445$\pm$0.097 \\
AT         &  0.793$\pm$0.071 &  0.502$\pm$0.082 &  0.884$\pm$0.012 &  0.417$\pm$0.062 \\
HDA         &  0.805$\pm$0.050 &  0.523$\pm$0.073 &  0.884$\pm$0.019 &  0.455$\pm$0.073 \\
\tname      &  {\bf 0.808$\pm$0.066} &  {\bf 0.535$\pm$0.061} &  {\bf 0.895$\pm$0.016} &   {\bf 0.460$\pm$0.076} \\ 
\hline
Rich Model &  0.854$\pm$0.069 &  0.657$\pm$0.077 &  0.922$\pm$0.014 &   0.595$\pm$0.070 \\
\bottomrule\vspace{-0mm}
\end{tabular}
}
\label{tb:compare_eeg}
\end{table}

\begin{table}[h!]
\centering
\caption{The $p$-values of corresponding $t$-tests (on one-tail) for every two samples of ROC-AUC scores of \mname and a tested benchmark (i.e., Direct, KD, AT and HDA) on MIMIC-III, PTBDB and EEG datasets, respectively. The corresponding significance percentage ($s$) is provided in the parentheses next to each reported $p$-value.}
\begin{tabular}{l|lll}
\toprule
         & MIMIC-III & PTBDB  & EEG    \\
\midrule
\hspace{-2mm}with Direct & 0.0000 (s = 01\%)    & 0.0154 (s = 05\%) & 0.1720 (s = 20\%) \\
\hspace{-2mm}with KD     & 0.0450 (s = 05\%)    & 0.1874 (s = 20\%) &  0.0124 (s = 05\%) \\
\hspace{-2mm}with AT     & 0.0007 (s = 01\%)    & 0.0421 (s = 05\%) &  0.1821 (s = 20\%)\\
\hspace{-2mm}with HDA    & 0.0000 (s = 01\%)    & 0.0741 (s = 10\%) &  0.4823 (s = 20\%) \\
\bottomrule
\end{tabular}
\label{tb:p_value}
\end{table}

\begin{figure*}[ht]
\centering
\begin{tabular}{ccc}
\hspace{-4mm}\includegraphics[width=5.8cm]{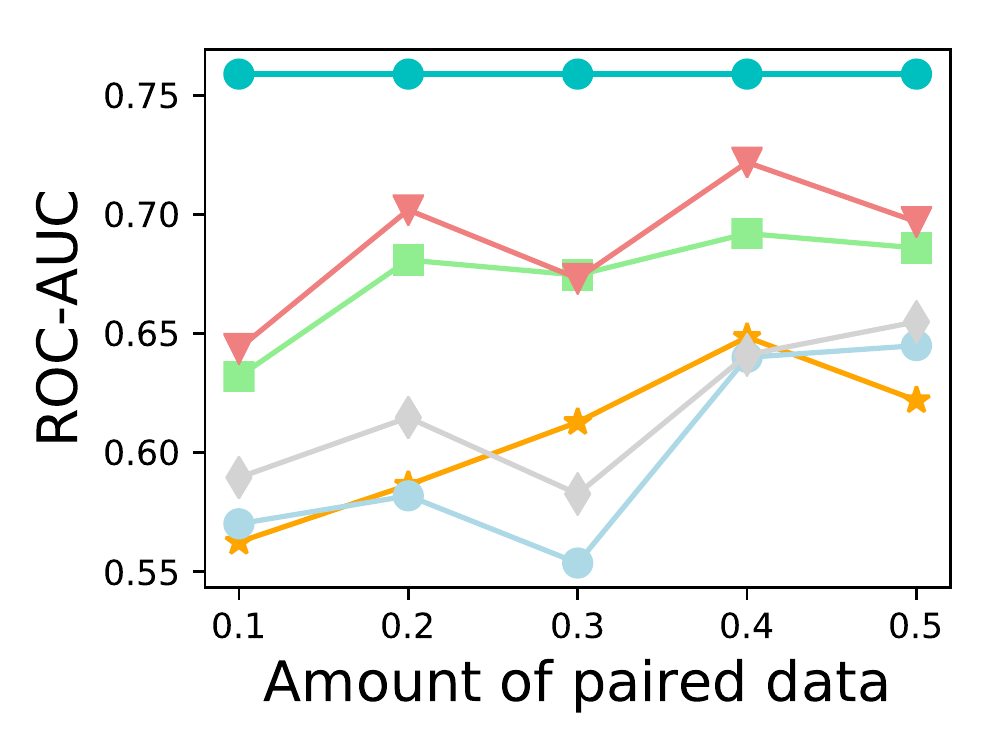} & 
\hspace{-0mm}\includegraphics[width=5.8cm]{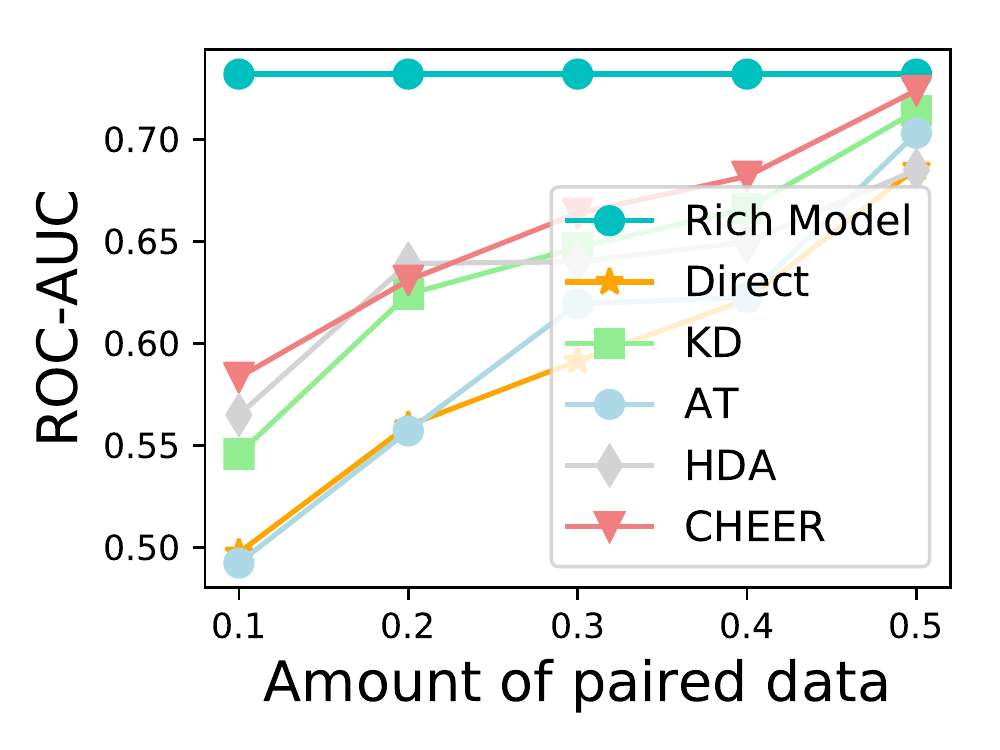} &
\hspace{-0mm}\includegraphics[width=5.8cm]{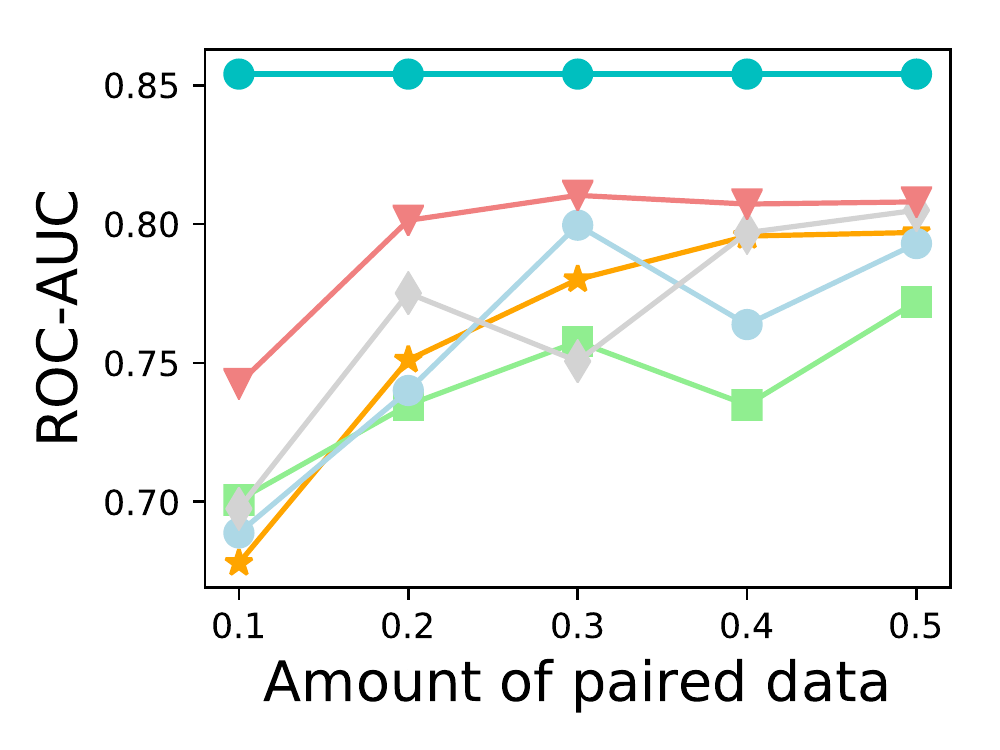}\\
\hspace{0mm} (a) MIMIC-III & 
\hspace{0mm} (b) PTBDB  &
\hspace{0mm} (c) EEG  \vspace{-2mm}\\
\end{tabular}    
\caption{Graphs of achieved ROC-AUC scores on (a) MIMIC-III, (b) PTBDB and (c) EEG of the infused models generated by Direct, KD, AT, HDA and \tname with different sizes of the paired datasets. The X-axis shows the ratio between the size of the paired dataset and that of the \emph{rich} dataset.\vspace{-3mm}}
\label{fig:varying1}
\end{figure*}

\begin{figure*}[ht]
\centering
\begin{tabular}{ccc}
\hspace{-4mm}\includegraphics[width=5.8cm]{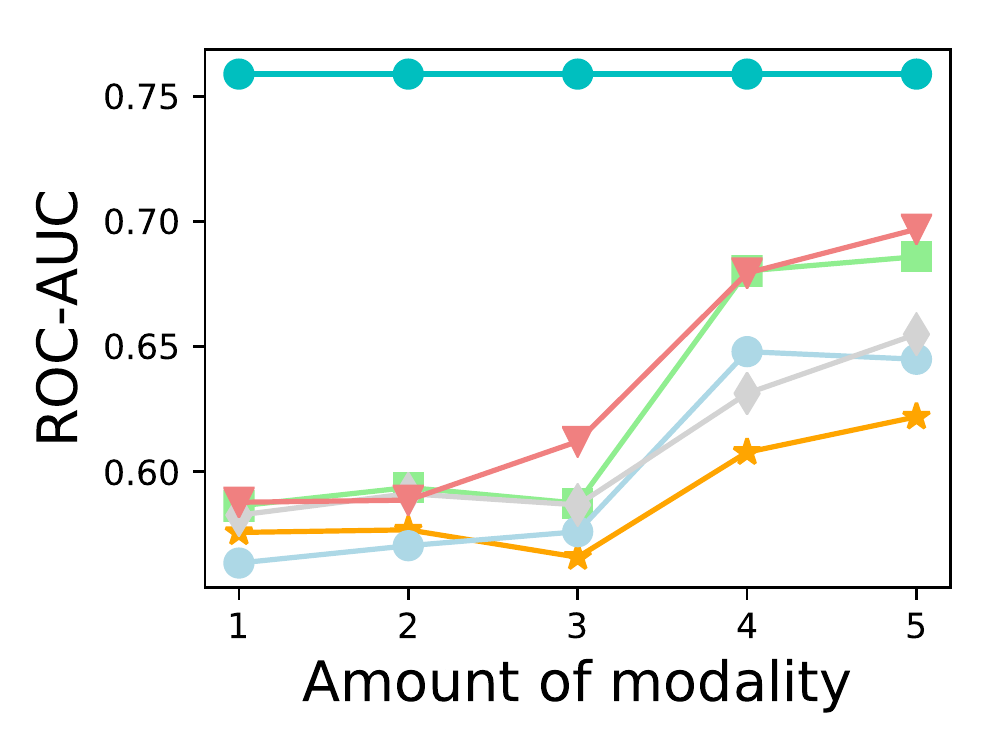} & 
\hspace{-0mm}\includegraphics[width=5.8cm]{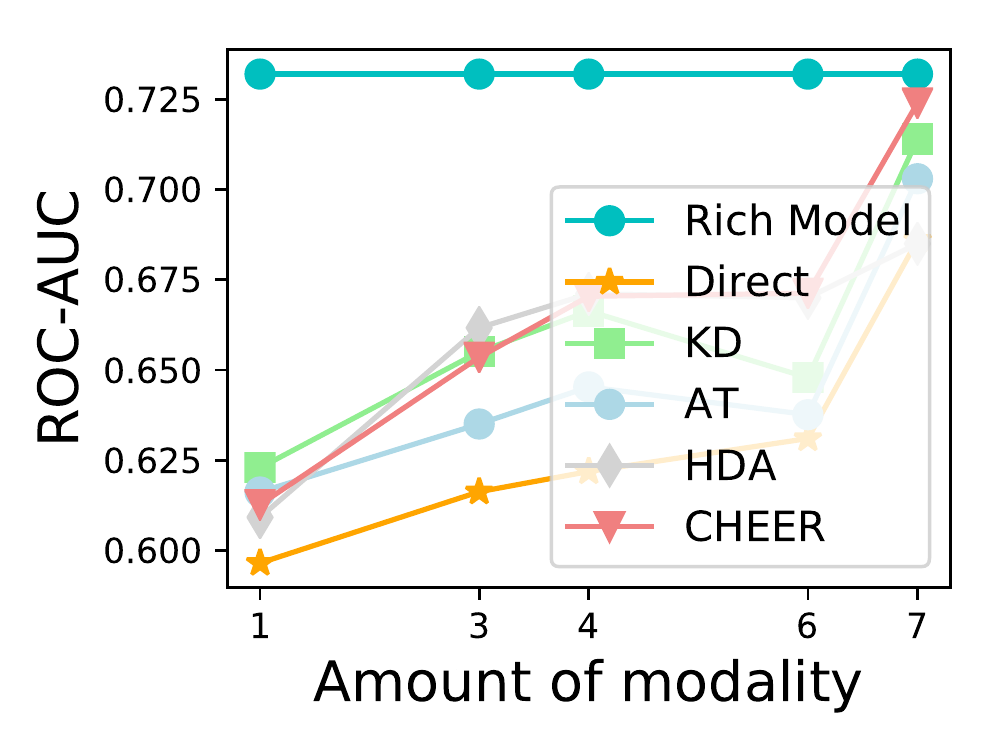} &
\hspace{-0mm}\includegraphics[width=5.8cm]{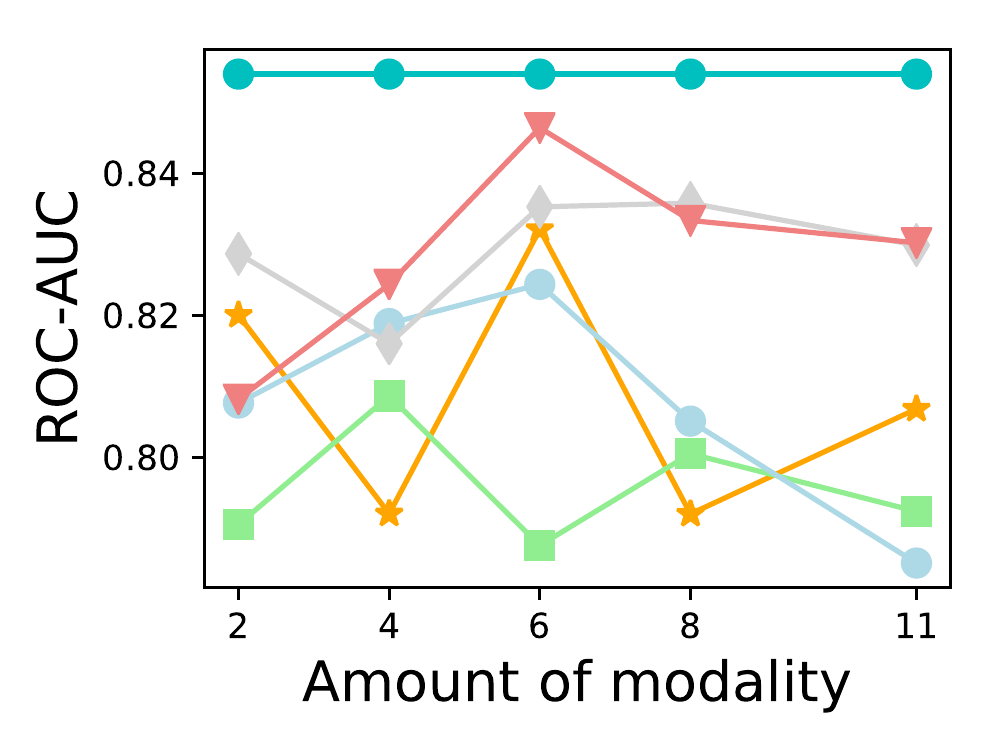}\\
\hspace{0mm} (a) MIMIC-III & 
\hspace{0mm} (b) PTBDB  &
\hspace{0mm} (c) EEG  \vspace{-2mm}\\
\end{tabular}    
\caption{Graphs of achieved ROC-AUC scores on (a) MIMIC-III, (b) PTBDB and (c) EEG of the infused models generated by Direct, KD, AT, HDA and \tname with different number of data channels (i.e., features) included in the poor dataset. Notice that all method use the same set of selected features for each run. \vspace{-0mm}}
\label{fig:varying2}
\end{figure*}

\subsection{Analyzing Knowledge Infusion Effect in Different Data Settings}~\label{sec:analysis}\vspace{-2mm}

\noindent To further analyze the advantages of \tname's knowledge infusion over those of the existing works (e.g., KD and AT), we perform additional experiments to examine how the variations in (1) sizes of the paired dataset and (2) the number of features of the \emph{poor} dataset will affect the infused model's performance. The results are shown in Fig.~\ref{fig:varying1} and Fig.~\ref{fig:varying2}, respectively. In particular, Fig.~\ref{fig:varying1} shows how the ROC-AUC of the infused model generated by each tested method varies when we increase the ratio between the size of the paired dataset and that of the \emph{rich} data. Fig.~\ref{fig:varying2}, on the other hand, shows how the infused model's ROC-AUC varies when we increase the number of features of the \emph{poor} dataset. In both settings, the reported performance of all methods is averaged over $10$ independent runs.\\

\noindent\textbf{Varying Paired Data.} Fig.~\ref{fig:varying1} shows that (a) \tname outperforms all baselines with varying sizes of the paired data and (b) direct learning on \emph{poor data} yields significantly worse performance across all settings. Both of which are consistent with our observations earlier on the superior knowledge infusion performance of \tname. 
The infused models generated by KD, HDA and AT both perform consistently worse than that of \tname by a substantial margin across all datasets. Their performance also fluctuates over a much wider range (especially on EEG data) than that of \tname when we vary the size of the paired datasets. This shows that \tname's knowledge infusion is more data efficient and robust under different data settings.\\

\noindent On another note, we also notice that when the amount of paired data increases from $20\%$ to $30\%$ of the \emph{rich} data, there is a performance drop that happens to all tested methods with attention transfer (i.e., \tname and AT) on MIMIC-III but not on PTBDB and EEG. This is, however, not surprising since unlike PTBDB and EEG, MIMIC-III comprises of more heterogeneous types of signals and its data distribution is also more unbalanced, which affects the attention learning, and causes similar performance drop patterns between methods with attention transfer such as \tname and AT.\\ 

\noindent\textbf{Varying The Number of Features.} Fig.~\ref{fig:varying2} shows how the prediction performance of the infused models generated by tested methods changes as we vary the number of features in \emph{poor data}. In particular, it can be observed that the performance of \tname's infused model on all datasets increases steadily as we increase the number of input features observed by the \emph{poor model}, which is expected.\\

\noindent On the other hand, it is perhaps surprising that as the number of features increases, the performance of KD, HDA, AT and Direct fluctuates more widely on PTBDB and EEG datasets, which is in contrast to our observation of \tname. This is, however, not unexpected since the informativeness of different features are different and hence, to utilize and combine them effectively, we need an accurate feature weighting/scoring mechanism. This is not possible in the cases of Direct, KD, HDA and AT because (a) Direct completely lacks knowledge infusion from the \emph{rich model}, (b) KD and HDA only performs target transfer from the \emph{rich} to \emph{poor} model, and ignores the weighting/scoring mechanism, and (c) AT only transfers the scoring mechanism to the \emph{poor} model (i.e., attention transfer) but not the feature aggregation mechanism, which is also necessary to combine the weighted features correctly. In contrast, \tname transfers both the weighting/scoring (via behavior infusion) and feature aggregation (via target infusion) mechanisms, thus performs more robustly and is able to produce steady gain (without radical fluctuations) in term of performance when the number of features increases. This supports our observations earlier regarding the lowest performance variance achieved by the infused model of \tname, which also suggests that \tname's knowledge infusion scheme is more robust than those of KD, HDA and AT.\\

\begin{table}[]
\centering
\caption{CHEER's performance on MIMIC-III, PTBDB and EEG with (left-column) $K$ features with highest mutual information (MI) to the class label as features of the poor dataset; and (right-column) $K$ features with lowest mutual information to the class label as features of the poor dataset. $K$ is set to 2 for MIMIC-III, 5 for PTBDB and 7 for EEG.}
\begin{tabular}{lcc}
\toprule
Dataset   & Highest MI Features      & Lowest MI Features   \\
\midrule
MIMIC-III & 0.672 $\pm$ 0.044 & 0.657 $\pm$ 0.012 \\
PTBDB     & 0.646 $\pm$ 0.133 & 0.639 $\pm$ 0.115 \\
EEG       & 0.815 $\pm$ 0.064 & 0.807 $\pm$ 0.042 \\
\bottomrule
\end{tabular}
\label{tb:mutual}
\end{table}

\noindent Finally, to demonstrate how the performance of CHEER varies with different choices of feature sets for poor data, we computed the mutual information between each feature and the class label, and then ranked them in decreasing order. The performance of CHEER on all datasets is then reported in two cases, which include (a) $K$ features with highest mutual information, and (b) $K$ features with lowest mutual information. In particular, the reported results (see Table~\ref{tb:mutual}) show that a feature set with low mutual information to the class label will induce worse transfer performance and conversely, a feature set (with the same number of features) with high mutual information will likely improve the transfer performance.\\

\noindent To further inspect the effects of used modalities in \tname, we also computed the averaged entropy of each modality across all classes, and ranked them in decreasing order for each dataset. Then, we selected a small number of top-ranked, middle-ranked and bottom-ranked features from the entire set of modalities. These are marked as Top, Middle and Bottom respectively in Table~\ref{tb:rank_modalities}.\\

\noindent The number of selected features for each rank is 2, 4 and 5 for MIMIC-III, PTBDB and EEG, respectively. Finally, we report the ROC-AUC scores achieved by the corresponding infused models generated by \tname for each of those feature settings in Table~\ref{tb:rank_modalities}. It can be observed from this table that the ROC-AUC of the infused model degrades consistently across all datasets when we change the features of \emph{poor} data from those in Top to Middle and then to Bottom. This verifies our statement earlier that the informativeness of different data features are different.

\begin{table}[h!]
\caption{\tname's performance using different sets of data features with different information quality (as measured by their entropy scores).}
\resizebox{0.95\columnwidth}{!}{
\begin{tabular}{llll}
\toprule
{} & MIMIC-III & PTBDB & EEG  \\
\midrule
Top & 0.688 $\pm$ 0.010 & 0.710 $\pm$ 0.131 & 0.839 $\pm$ 0.044 \\
Middle & 0.676 $\pm$ 0.014 & 0.682 $\pm$ 0.132 & 0.788 $\pm$ 0.065 \\
Bottom & 0.664 $\pm$ 0.012 & 0.633 $\pm$ 0.130 & 0.758 $\pm$ 0.066 \\
\hline
Rich & 0.759 $\pm$ 0.014 & 0.732 $\pm$ 0.110 & 0.854 $\pm$ 0.069 \\
\bottomrule
\end{tabular}
}
\label{tb:rank_modalities}
\end{table}

\section{Conclusion}
\label{conclusion}
This paper develops a knowledge infusion framework (named \mname) that helps infuse knowledge acquired by a \emph{rich model} trained on feature-rich data with a \emph{poor model} which only has access to feature-poor data. The developed framework leverages a new model representation to re-parameterize the \emph{rich model} and consequently, consolidate its learning behaviors into succinct summaries that can be infused efficiently with the \emph{poor model} to improve its performance. To demonstrate the efficiency of \mname, we evaluated \mname on multiple real-world datasets, which show very promising results. We also develop a formal theoretical analysis to guarantee the performance of \mname under practical assumptions. Future extensions of \mname includes the following potential settings: incorporating meta/contextual information as part of the features and/or learning from data with missing labels.

\section{Acknowledgments}
This work is part supported by National Science Foundation award IIS-1418511, CCF-1533768 and IIS-1838042, the National Institute of Health award NIH R01 1R01NS107291-01 and R56HL138415.

\bibliographystyle{ACM-Reference-Format}
\bibliography{references}

\end{document}